\PassOptionsToPackage{dvipsnames}{xcolor}
\documentclass[]{fairmeta}
\usepackage{amssymb}
\usepackage{amsmath}
\usepackage{multirow}
\usepackage{todonotes}
\usepackage{wrapfig}
\usepackage{url}
\usepackage[
  style=numeric-comp,
  sorting=none,
  maxbibnames=99,
  minbibnames=99,
  maxcitenames=2,
  mincitenames=1,
  giveninits=true,
  terseinits=true,
  doi=false,
  isbn=false,
  url=false,
  eprint=false
]{biblatex}
\DeclareNameAlias{sortname}{family-given}
\DeclareNameAlias{default}{family-given}
\usepackage{xcolor}
\usepackage{dirtree}
\DeclareFieldFormat[article,inproceedings,incollection]{title}{{#1}}
\DeclareFieldFormat{booktitle}{#1}
\DeclareFieldFormat{journaltitle}{#1}
\renewbibmacro{in:}{}
\AtEveryBibitem{%
  \clearfield{pages}%
  \clearfield{volume}%
  \clearfield{number}%
  \clearfield{month}%
  \clearlist{organization}%
  \clearlist{publisher}%
  \clearlist{location}%
  \clearfield{series}%
  \clearfield{note}%
}
\renewbibmacro*{issue+date}{%
  \printtext[parens]{\printfield{year}}%
  \newunit}
\renewbibmacro*{date}{%
  \printtext[parens]{\printfield{year}}}

\usepackage{graphicx}
\usepackage{float}
\usepackage{booktabs}
\usepackage{siunitx}
\usepackage{makecell}
\usepackage{placeins}
\usepackage{enumitem}
\usepackage{xspace}

\usepackage{listings}

\definecolor{codebg}{RGB}{245,245,245}
\definecolor{codekey}{RGB}{0,0,180}       %
\definecolor{codestr}{RGB}{163,21,21}     %
\definecolor{codecom}{RGB}{0,128,0}       %
\definecolor{codenum}{RGB}{128,0,128}     %
\definecolor{codeframe}{RGB}{200,200,200}

\definecolor{recgenorange}{RGB}{224,208,245}  %
\definecolor{turquoise}{cmyk}{0.65,0,0.1,0.3}
\definecolor{purple}{rgb}{0.65,0,0.65}
\definecolor{dark_green}{rgb}{0, 0.5, 0}
\definecolor{orange}{rgb}{0.8, 0.6, 0.2}
\definecolor{red}{rgb}{0.8, 0.2, 0.2}
\definecolor{darkred}{rgb}{0.6, 0.1, 0.05}
\definecolor{blueish}{rgb}{0.3, 0.3, .6}
\definecolor{light_gray}{rgb}{0.7, 0.7, .7}
\definecolor{pink}{rgb}{1, 0, 1}
\definecolor{greyblue}{rgb}{0.25, 0.25, 1}
\definecolor{awesome}{rgb}{1.0, 0.13, 0.32}

\definecolor{neonviolet}{RGB}{168, 85, 247}     %
\definecolor{accentdark}{RGB}{74, 29, 142}      %

\lstdefinestyle{codebase}{
  basicstyle=\ttfamily\small,
  backgroundcolor=\color{codebg},
  frame=single,
  rulecolor=\color{codeframe},
  framesep=4pt,
  breaklines=true,
  breakatwhitespace=true,
  columns=fullflexible,
  keepspaces=true,
  showstringspaces=false,
  tabsize=2,
  captionpos=b,
  keywordstyle=\color{codekey}\bfseries,
  stringstyle=\color{codestr},
  commentstyle=\color{codecom}\itshape,
}

\lstdefinelanguage{json}{
  morestring=[b]",
  morestring=[d]',
  stringstyle=\color{codestr},
  literate=
    *{:}{{{\color{codekey}{:}}}}{1}
     {,}{{{\color{codekey}{,}}}}{1}
     {\{}{{{\color{codekey}{\{}}}}{1}
     {\}}{{{\color{codekey}{\}}}}}{1}
     {[}{{{\color{codekey}{[}}}}{1}
     {]}{{{\color{codekey}{]}}}}{1},
}

\lstdefinelanguage{yaml}{
  keywords={true,false,null,True,False,Null,yes,no,on,off},
  keywordstyle=\color{codekey}\bfseries,
  sensitive=true,
  comment=[l]{\#},
  commentstyle=\color{codecom}\itshape,
  morestring=[b]',
  morestring=[b]",
  stringstyle=\color{codestr},
  literate=
    *{:}{{{\color{codekey}{:}}}}{1}
     {-}{{{\color{codekey}{-}}}}{1},
}

\makeatletter
\newcommand\footnoteref[1]{\protected@xdef\@thefnmark{\ref{#1}}\@footnotemark}
\makeatother

\newcommand{\TODO}[1]{}
\newcommand{\method}{\textsc{RecGen}\xspace}

\newcommand{\objectpose}{\mathbf{T}}
\newcommand{\myparagraph}[1]{\paragraph{#1}}

\newcommand{\image}{\mathbf{I}}
\newcommand{\depth}{\mathbf{D}}
\newcommand{\pointmap}{\mathbf{P}}
\newcommand{\mask}{\mathbf{M}}
\newcommand{\imagedim}{d}

\newcommand{\shape}{\mathbf{s}}
\newcommand{\appearance}{\mathbf{a}}

\newcommand{\sz}[1]{}
\newcommand{\zi}[1]{}
\newcommand{\az}[1]{}
\newcommand{\zw}[1]{}
\newcommand{\lb}[1]{}
\newcommand{\rahaf}[1]{}

\newcommand{\Fig}[1]{\hyperref[{#1}]{Figure~\ref*{#1}}}
\newcommand{\fig}[1]{\hyperref[{#1}]{Fig.~\ref*{#1}}}

\newcommand{\Tab}[1]{\hyperref[{#1}]{Table~\ref*{#1}}}
\newcommand{\tab}[1]{\hyperref[{#1}]{Table~\ref*{#1}}}
\newcommand{\Eqn}[1]{\hyperref[{#1}]{Equation~\ref*{#1}}}
\newcommand{\eqn}[1]{\hyperref[{#1}]{Eq.~\ref*{#1}}}

\newcommand{\sect}[1]{\hyperref[{#1}]{Sec.~\ref*{#1}}}
\newcommand{\supp}[1]{\hyperref[{#1}]{Suppl.~\ref*{#1}}}
\newcommand{\app}[1]{\hyperref[{#1}]{App.~\ref*{#1}}}
\newcommand{\App}[1]{\hyperref[{#1}]{Appendix~\ref*{#1}}}

\newcommand{\maketitlesupplementary}{%
  \newpage
  \begin{center}
    {\huge\sffamily\bfseries Supplementary Material}\\[1em]
  \end{center}
}

\title{Reconstruction by Generation: 3D Multi-Object Scene Reconstruction from Sparse Observations}

\author[1*]{Andrii Zadaianchuk}
\author[1*]{Leonardo Barcellona}
\author[1]{Lennard Schuenemann}
\author[1]{Christian Gumbsch}
\author[2]{Zehao Wang}
\author[4]{Muhammad Zubair Irshad}
\author[3]{Fabien Despinoy}
\author[3]{Rahaf Aljundi}
\author[1]{Stratis Gavves}
\author[4*\dagger]{Sergey Zakharov}

\affiliation[1]{University of Amsterdam}
\affiliation[2]{KU Leuven}
\affiliation[3]{Toyota Motor Europe}
\affiliation[4]{Toyota Research Institute}

\contribution[*]{Core Contributor}
\contribution[\dagger]{Project Lead}

\abstract{%
Accurately reconstructing complex full multi-object scenes from sparse observations remains a core challenge in computer vision and a key step toward scalable and reliable simulation for robotics. In this work, we introduce \method, a generative framework for probabilistic joint estimation of object and part shapes, as well as their pose under occlusion and partial visibility from one or multiple RGB-D images. By leveraging compositional synthetic scene generation and strong 3D shape priors, \method generalizes across diverse object types and real-world environments. \method achieves state-of-the-art performance on complex, heavily occluded datasets, robustly handling severe occlusions, symmetric objects, object parts, and intricate geometry and texture. Despite using nearly 80\% fewer training meshes than the previous state of the art SAM3D, \method outperforms it by 30.1\% in geometric shape quality, 9.1\% in texture reconstruction, and 33.9\% in pose estimation. 
}

\metadata[Project page]{\href{https://reconstruction-by-generation.github.io}{reconstruction-by-generation.github.io}}

\begin{document}
\maketitle

\begin{center}
    \includegraphics[width=\columnwidth]{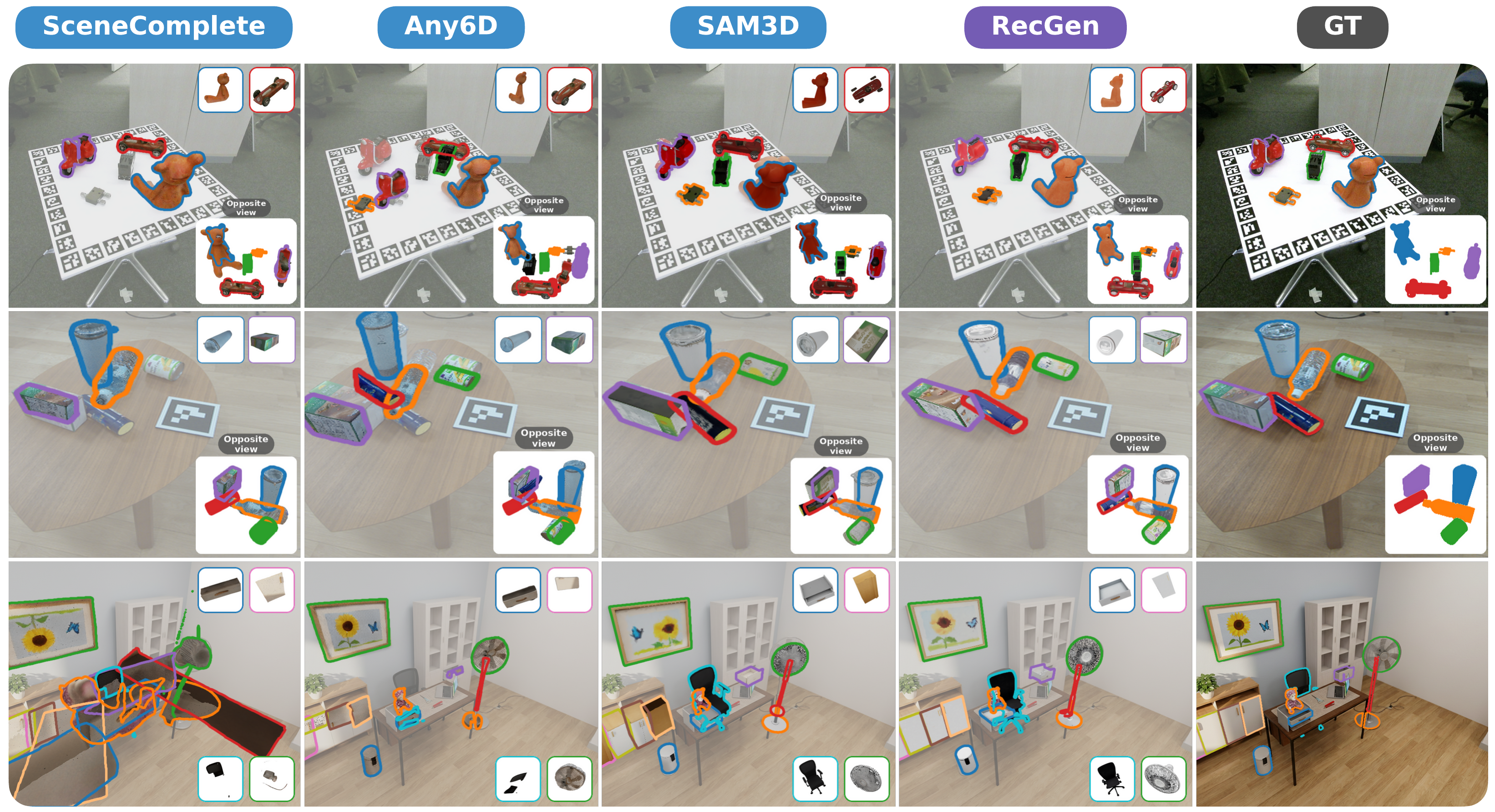}
    \captionsetup{width=\columnwidth}
    \vspace{-5mm}
    \captionof{figure}{
    \emph{\method} generates full reconstructions of complex occluded scenes from single or multiple RGB-D images, enabling robust generation of digital twin replicas of real-world environments.Our model recovers occluded geometry of both objects and parts, is robust to imperfect sensor depth, and handles object symmetries — challenges that most recent baselines struggle to address.\looseness=-1
    \label{fig:teaser}
    \vspace{-3mm}
    }
\end{center}

\section{Introduction}
\label{sec:intro}
Simulation is increasingly used to train and evaluate embodied AI systems~\cite{li2023behavior, savva2019habitat,mittal2025isaac, chen2025robotwin}, however, its overall impact is limited by the substantial cost and complexity associated with constructing high-fidelity digital twins~\cite{tao2024advancements}. Constructing such twins typically requires detailed scanning and manual registration of objects within scenes, a labor-intensive process that is difficult to scale.
A promising alternative is to recover structured multi-object scenes directly from sparse observations~\cite{zhu2024living,chen2025sam}.
However, accurately estimating object geometry and 6-DoF pose from limited RGB-D input in cluttered environments remains fundamentally challenging.
Occlusions, object symmetries, complex geometry, and noisy depth observations make pose estimation under partial visibility brittle, posing challenges for scalable real-to-sim reconstruction\cite{ikeda2024diffusionnocs}.

Generative 3D models~\cite{liu2023zero,xiang2025structured,yang2024hunyuan3d}
have recently demonstrated strong potential for reconstructing real-world objects from sparse observations. In parallel, model-free pose estimation methods~\cite{lee2025any6d,liu2022gen6d,agarwal2024scenecomplete,wen2024foundationpose} leverage generated 3D shapes to perform pose registration against input images or depth maps. However, these approaches treat shape generation, completion, and pose estimation as separate stages, increasing complexity, compounding errors, and reducing robustness under occlusion. In contrast, we propose \method, a unified generative framework that jointly infers object geometry and 6-DoF pose from single or multiple RGB-D observations, enabling coherent reasoning about shape and pose under uncertainty. The novel design and training recipe of \method overcomes central limitations faced by existing 3D reconstruction methods, as qualitatively illustrated in \fig{fig:teaser}.\looseness=-1

The first limitation lies in pose registration where shapes predicted by image-conditioned generative models~\cite{xiang2025structured, xu2024instantmesh} must be aligned post hoc to observed RGB or depth data, a process that is often brittle in cluttered scenes and for symmetric or weakly textured objects. In contrast, \method performs joint probabilistic estimation of shape and pose directly in the camera frame, enabling geometrically consistent object reconstruction without requiring separate registration stages.

Related to this problem is object completion under partial visibility where existing generative models either misinterpret visible regions as complete geometry or fail to infer the occluded or unobserved regions due to limited contextual cues. This issue largely arises from training on occlusion‑free objects or masked images where the target object is fully visible, unlike real‑world conditions.
To address this issue, we introduce a large-scale synthetic dataset of occluded objects, which enables \method to learn priors that support robust shape completion under challenging occlusions.  
Importantly, unlike prior methods that take masked images as direct input, we encode masks as positional signals indicating which pixels belong to the object of interest, providing richer contextual information for reasoning about occlusions.

Symmetry further complicates pose estimation and texture reconstruction.
The 6-DoF pose estimation of symmetric objects is inherently ambiguous. For objects such as bottles or boxes with semantic labels, textures must respect the object-to-camera orientation to correctly place view-dependent details. Without explicit pose conditioning, texturing networks often produce inconsistent or misaligned textures, as observed in recent generative methods such as SAM3D~\cite{chen2025sam}.
To overcome this challenge, we explicitly condition texture reconstruction in \method on the estimated object pose, enabling view-consistent and semantically aligned texturing even in the presence of geometric symmetries. 
 Another limitation is that existing methods reconstruct objects as single monolithic meshes, without recovering their internal part structure~\cite{xu2024instantmesh,xiang2025structured}. 
However, estimating such shapes and poses is essential to learn part-level control tasks such as articulated object manipulation~\cite{yu2025artgs,jiang2025dexsim2real}.
To address this, \method supports part-level shape and pose estimation by unifying scene decomposition into objects, and objects into parts within a single generalizable framework, by extending our object-level training with part-annotated assets.

\setlength{\parskip}{0pt}
Whereas depth images provide geometric cues that improve both shape and pose estimation,  most generative models remain RGB-centric and use depth only in alignment stages~\cite{xu2024instantmesh, xiang2025structured}, resulting in error-prone multi-step pipelines.
Although SAM3D~\cite{chen2025sam} supports depth input, it is sensitive to commodity sensor noise, leading to pose misalignment and degraded reconstruction. We address this limitation by training on realistically estimated depth from FoundationStereo~\cite{wen2025foundationstereo}, rather than relying on perfect rendered depth%
, enabling  \method to leverage 3D structural cues while remaining robust to imperfect measurements. Finally, an important capability largely unsupported by current generative models~\cite{chen2025sam, xiang2025structured, xu2024instantmesh} is multi-view conditioning, despite its practical relevance in real-world setups where multiple cameras are often available. We train \method to explicitly support both single-view and multi-view conditioning within a unified framework. In the multi-view setting, the model can integrate complementary observations across views, reduce reconstruction ambiguities, and improve both geometric consistency and pose accuracy. This enables \method to better exploit additional visual information when available.
\begin{figure*}[!t]
    \centering
    \includegraphics[width=\linewidth]{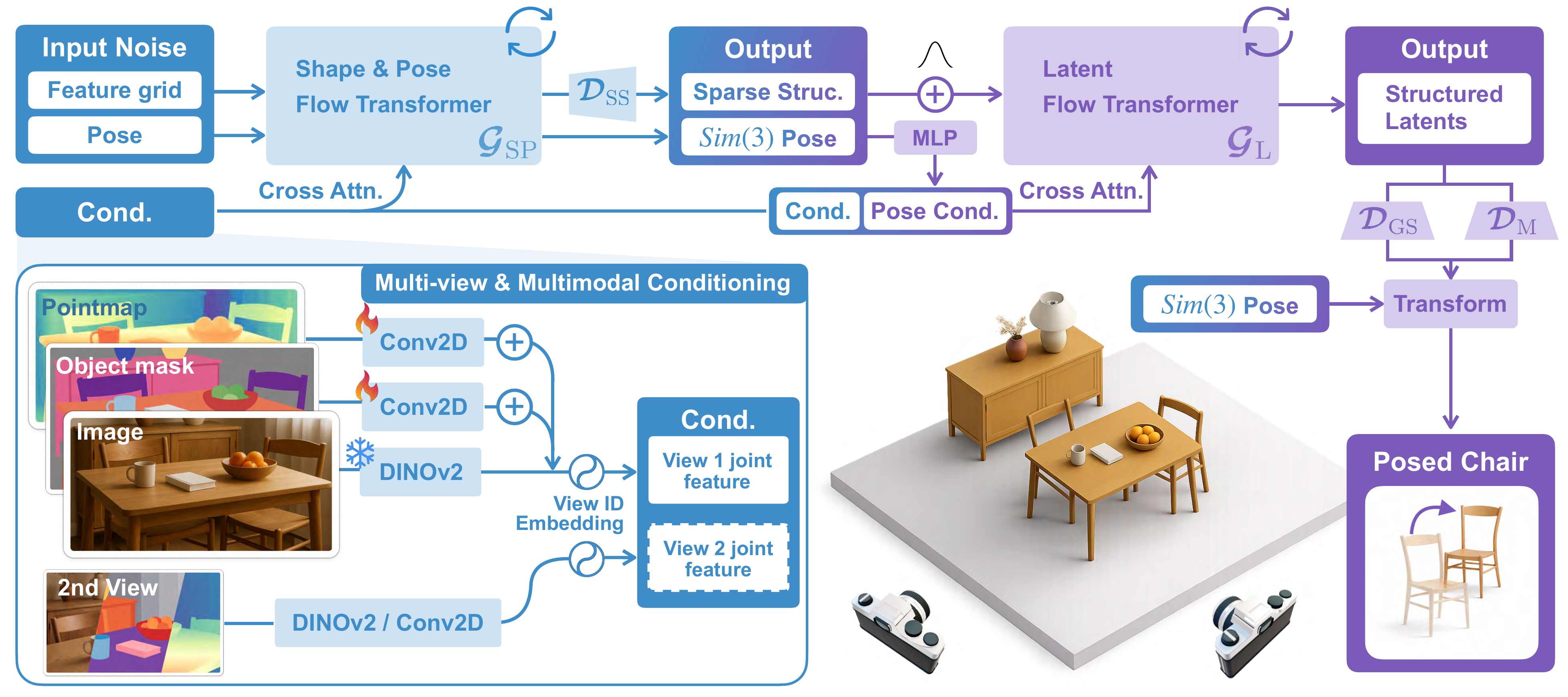}
    \caption{\textbf{The \method architecture}. Given one or more input observations consisting of RGB images, Point maps and Object masks, our framework (1) predicts a sparse object structure and its pose in a normalized camera frame, and (2) recovers textured meshes. Both stages employ flow transformer models conditioned on multimodal features, together with dedicated decoders to recover sparse structure, mesh, and texture.
    }

    \label{fig:method_overview}
\end{figure*}

\setlength{\parskip}{0pt}
To this end, \method is a novel framework and training recipe for joint shape completion and pose estimation that bridges the gap between generative modeling and real-world 3D reconstruction. %
Our main contributions are:
\begin{itemize}[label=\textbullet,leftmargin=1.2em,itemsep=0pt,topsep=1pt,parsep=0pt]
    \item We propose \method, a multi-hypothesis framework for jointly estimating object pose and complete 3D textured shape from one or a few images, without any prior knowledge of the object.
    \item We introduce a synthetic dataset of occluded objects and their parts, enabling robust RGB-D training under heavy occlusion and object symmetries.
    \item \method sets a new state of the art, surpassing SAM3D by 30.1\% in geometric shape quality, 9.1\% in texture reconstruction, and 33.9\% in pose estimation, while trained on 80\% less data.
    \item Extensive experiments show that \method generalizes robustly to part-level, occluded, symmetric, and uncommon objects.
\end{itemize}

\section{Related Work}
\label{sec:related_work}

\subsection{Pose and shape prediction}

Joint modeling of object shape and pose in the camera frame has emerged as an important direction for scene-level 3D reconstruction, supported by rapid progress in the two problems independently. On the shape side, image-conditioned 3D generation has advanced with feed-forward and hybrid generative models such as CRM~\cite{wang2024crmsingleimage3d}, LGM~\cite{tang2024lgmlargemultiviewgaussian}, InstantMesh~\cite{xu2024instantmesh}, TRELLIS~\cite{xiang2025structured}, and Hunyuan3D~\cite{hunyuan3d22025tencent}, which improve fidelity and spatial consistency via stronger geometric and latent priors. In parallel, methods for novel-object 6D pose estimation have also achieved strong performance.
FoundationPose~\cite{wen2024foundationpose} unifies model-based and model-free 6D pose estimation and tracking for novel objects, while Any6D~\cite{lee2025any6d} focuses on model-free pose estimation from a single RGB-D anchor observation. In practice, shape and pose are jointly required and geometrically coupled, motivating recent efforts toward joint prediction.
Approaches to joint shape–pose reconstruction fall into two categories: modular pipelines and unified feed-forward methods. Modular approaches (e.g., GigaPose~\cite{nguyen2024gigapose}, Pos3R~\cite{Pos3R}, OmniShape~\cite{liu2025omnishapezeroshotmultihypothesisshape}, SceneComplete~\cite{schonberger2016structure}, Gen3DSR~\cite{ardelean2025gen3dsrgeneralizable3dscene}) typically combine an image-to-3D reconstructor~\cite{xu2024instantmesh} with a separate pose alignment stage based on correspondences, depth, or registration. While flexible, such pipelines decouple geometry and pose and may suffer from error propagation. In contrast, unified approaches predict both in a single forward pass. Methods such as CenterSnap~\cite{irshad2022centersnap} and ShAPO~\cite{irshad2022shapo} jointly estimate complete 3D shape and 6D pose using camera-centered spatial representations.
Recently, scene-level generative methods have started to jointly model instance geometry and spatial arrangement. MIDI \cite{li2025midi} performs multi-instance diffusion to generate coherent 3D scenes from a single image. SAM3D~\cite{chen2025sam} tackles generative monocular reconstruction, predicting object geometry together with scene layout in the camera frame. These approaches are largely formulated under a single monocular condition and do not naturally support richer multi-view conditioning.
Recent works have also explored part-level 3D generation, synthesizing objects as collections of semantic components rather than monolithic meshes~\cite{chen2025partgen,he2026unipart,zhang2025bang,lin2025partcrafter}. However, they typically focus on part decomposition or synthesis rather than pose reasoning from observations.
Our approach enables joint shape and pose estimation across multiple conditions, supporting diverse inputs such as RGB-D and multi-view observations while inferring both object- and part-level representations.

\subsection{Real-to-Sim in Robotics}

Scene reconstruction and generation has gathered significant interest in robotics~\cite{melnik2026digitaltwingenerationvisual, irshad2024neuralfieldsroboticssurvey}. Among emerging scene representations, 3D Gaussian Splatting~\cite{kerbl3Dgaussians} has gained attention due to its photorealistic rendering quality and explicit, point-based structure, which facilitates downstream robotics manipulation and navigation applications~\cite{yu2025pogs, qureshi2024splatsimzeroshotsim2realtransfer, shorinwa2024splat, abouchakra2024physicallyembodiedgaussiansplatting, ji2024-graspsplats, chhablani2025embodiedsplatpersonalizedrealtosimtorealnavigation, escontrela2025gaussgymopensourcerealtosimframework, shen2023distilled, yang2025noveldemonstrationgenerationgaussian, jiang2025gsworld}.

Building on these developments, recent robotics research underscores the importance of scene-level 3D generation in real-to-sim pipelines. Several approaches employ reconstructed or generated 3D scenes as intermediate representations for policy learning; for instance, X-Sim~\cite{dan2025x}, DreMa~\cite{barcellona2025dream}, Real2Render2Real~\cite{yu2025real2render2realscalingrobotdata}, and ZeroBot~\cite{zerobot2026} depend on such simulation assets for robot learning. Complementary work focuses on scaling policy evaluation, as in Real2Sim-Eval~\cite{zhang2026real2simeval}, RobotArena~\cite{jangir2025robotarenainftyscalablerobot}, and PolaRiS~\cite{jain2025polaris}, which transform real scenes into interactive simulation environments for reproducible benchmarking.
More recently, a separate line of work has focused on making generated scenes physically usable through post-hoc refinement, e.g., via physics-consistent inter-object reasoning or physics-aware joint shape-pose optimization in cluttered environments~\cite{yu2026picasso,xiang2026real,huang2026sim}. Taken together, these works suggest that robotics increasingly demands scene models that are not only visually faithful, but also accurate at the scene level, physically plausible, and readily usable under multi-view observations. Our method targets this need by providing a high-fidelity scene-level generative base model with support for multi-view conditioning.

\section{Method}
\label{sec:method}

\begin{figure*}[!t]
    \centering
    \includegraphics[width=\linewidth]{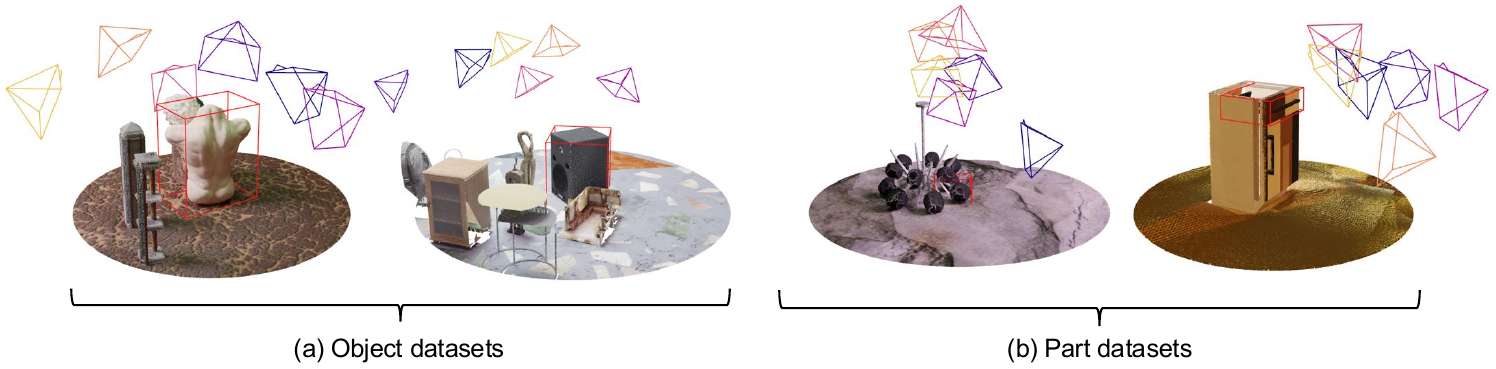}
    \caption{\textbf{RecGen Training Dataset Samples.} Representative examples of 3D assets from our training dataset, including compositional scenes with (a) objects from Objaverse-XL, ABO, HSSD, and (b) parts in object scenes the part-based datasets (PhysXNet, PartNext, PartNet-Mobility). Using such conditioning allows for scene-aware 3D generation of objects and their parts from partially occluded and posed objects, jointly with robust pose estimation of the corresponding assets.}
    \label{fig:dataset_samples}
\end{figure*}

Given one or two views of a real-world scene, we aim to reconstruct structured assets that serve as digital twins of the physical objects. Given $v$ as the view index, we assume that each input viewpoint provides an RGB image $\image^{(v)} \in \mathbb{R}^{\imagedim \times \imagedim \times 3}$, a depth map $\depth^{(v)} \in \mathbb{R}^{\imagedim \times \imagedim}$, camera intrinsics $\mathbf{K}^{(v)} \in \mathbb{R}^{3 \times 3}$, and a set of segmented regions $\mask^{(v)}$ corresponding to the observed objects or object parts. Our objective is to estimate the \emph{shape} $\shape$, \emph{pose} $\objectpose^{(v)}$, and \emph{appearance} $\appearance$ for each segmented region $\mask^{(v)}$, 
where $\objectpose^{(v)} \in \mathrm{Sim}(3)$ denotes a similarity transformation that maps object-centric coordinates to the normalized input frame.

Since an object's pose cannot be fully determined without knowledge of its shape, and vice versa, 
we jointly model these quantities rather than estimating them independently. 
Formally, we consider the joint conditional distribution
\[
p\big(\shape, \appearance, \{\mathbf{T}^{(v)}\}_v 
\mid 
\{\image^{(v)}, \depth^{(v)}, \mathbf{K}^{(v)}\}_v \big),
\]
which is highly complex and inherently multimodal.

To effectively model this distribution, we employ a generative framework based on rectified flow~\cite{lipman2023flow} 
that jointly produces high-quality 3D object shapes and their corresponding similarity transformations. 
\Fig{fig:method_overview} provides an overview of the proposed framework.

\subsection{Reconstruction by Generation}\label{sec:generation}

Our reconstruction framework (\fig{fig:method_overview}) consists of two main stages: (1) \emph{object structure and pose generation}, and (2) \emph{high-quality asset recovery}, both trained using rectified flow models to efficiently capture complex data distributions.

\subsubsection{Object Structure and Pose Generation.}
In the first stage, we jointly generate the object's sparse structure 
$\{\boldsymbol{p}_i\}_{i=1}^{L}$ together with its pose 
$\objectpose \in \mathrm{Sim}(3)$, parameterized by rotation 
$\boldsymbol{R} \in \mathrm{SO}(3)$, translation 
$\boldsymbol{t} \in \mathbb{R}^3$, and isotropic scale $s \in \mathbb{R}^+$. 
The transformation $\objectpose$ maps object-centric coordinates to the normalized input frame.
To enable dense tensor processing, the sparse voxel coordinates are converted into a dense binary occupancy grid 
$\boldsymbol{O} \in \{0,1\}^{64 \times 64 \times 64}$, where active voxels are set to $1$. The direct generation of $\boldsymbol{O}$ is computationally expensive. 
We therefore employ a 3D convolutional VAE to encode it into a lower-resolution continuous feature grid 
$\boldsymbol{S} \in \mathbb{R}^{16 \times 16 \times 16 \times 8}$, 
providing a smooth latent space suitable for rectified flow training with minimal information loss.

The pose $\objectpose$ is jointly denoised alongside $\boldsymbol{S}$ by concatenating its parameters with the structure features as an additional token, enabling the model to exploit geometric consistency between shape and pose. 
At each timestep $t$, the generator $\boldsymbol{\mathcal{G}}_{\mathrm{SP}}$ predicts velocity fields for both $\boldsymbol{S}$ and $\objectpose$, which are updated via Euler integration.

A transformer-based generator $\boldsymbol{\mathcal{G}}_{\mathrm{SP}}$ is trained to jointly produce $\boldsymbol{S}$ and $\objectpose$ from noisy inputs. 
The serialized input grid is augmented with positional encodings and processed by a transformer with adaptive layer normalization (AdaLN) and gating mechanisms~\cite{peebles2023scalable}. 
Conditioning is provided through cross-attention on our multimodal features formed by DINOv2~\cite{oquab2023dinov2} image features, as well as point map and mask features extracted from respective inputs.

The denoised feature grid $\boldsymbol{S}$ is decoded into the discrete occupancy grid $\boldsymbol{O}$ using a decoder $\boldsymbol{\mathcal{D}}_{\mathrm{SS}}$, of the same VAE, and converted back into active voxels $\{\boldsymbol{p}_i\}_{i=1}^{L}$, representing the predicted sparse object structure. 
The denoised transformation $\objectpose$ is applied to recover the object's rotation, translation, and scale.

\myparagraph{\textbf{Pose parameterization and normalization.}}
Inspired by \cite{geist2024learning}, we adopt pose parameterizations that avoid discontinuities, which could impair gradient-based optimization. In particular, we use the $6$D continuous representation proposed in~\cite{zhou2019continuity} as our main rotation representation for the structure generator $\boldsymbol{\mathcal{G}}_{\mathrm{SP}}$, which stores the first two columns of a rotation matrix and recovers the third via Gram--Schmidt orthogonalization. For the latent generator $\boldsymbol{\mathcal{G}}_{\mathrm{L}}$, we use the 9D pose parametrization as it was shown to perform best for model inputs.
Both representations are extended with a translation vector $\boldsymbol{t} \in \mathbb{R}^3$ and an isotropic scale $s \in \mathbb{R}^{+}$, yielding the full pose $\objectpose = \{\boldsymbol{R}, \boldsymbol{t}, s\}$. 

In addition to choosing an appropriate rotation representation, we apply $z$-score normalization to all pose components computed over the entire training set: \looseness=-1
\begin{equation*}\tilde{\objectpose} = \left\{(\boldsymbol{\rho}-\boldsymbol{\mu}_\rho)/\boldsymbol{\sigma}_\rho,\; (\boldsymbol{t}-\boldsymbol{\mu}_t)/\boldsymbol{\sigma}_t,\; (s-\mu_s)/\sigma_s \right\}, \end{equation*}
where $\boldsymbol{\rho}$ denotes the rotation parameters (quaternion or 6D), and $\boldsymbol{\mu}, \boldsymbol{\sigma}$ are component-wise means and standard deviations over the training dataset. This standardization ensures zero mean and unit variance for each component, preventing any single quantity from dominating the flow matching objective. During inference, we denormalize via \begin{equation*}\objectpose = \{\tilde{\boldsymbol{\rho}} \cdot \boldsymbol{\sigma}_\rho + \boldsymbol{\mu}_\rho, \tilde{\boldsymbol{t}} \cdot \boldsymbol{\sigma}_t + \boldsymbol{\mu}_t, \tilde{s} \cdot \sigma_s + \mu_s\}.\end{equation*}

\myparagraph{\textbf{Dynamic cropping and mask conditioning.}}
To specify the target object, most object-centric approaches apply segmentation masks to the RGB image, retaining only foreground RGB pixels. However, this discards contextual environment information that can help infer occlusions and scene layout. At the same time, providing the full image is both expensive and unnecessary, since most of the important context information is contained in the object's vicinity. Instead, we dynamically crop the original image and corresponding binary object mask to the region around the object during training, allowing for anywhere from 20\% to 100\% padding around the object. We encode the obtained mask $\mask \in \{0,1\}^{\imagedim \times \imagedim}$ using a learnable convolutional layer and inject the resulting feature map by adding it to the image features. This design allows the model to exploit both foreground and background cues when generating the object's structure and pose.~\looseness=-1

\begin{figure}[t]
    \centering
        \includegraphics[width=\linewidth]{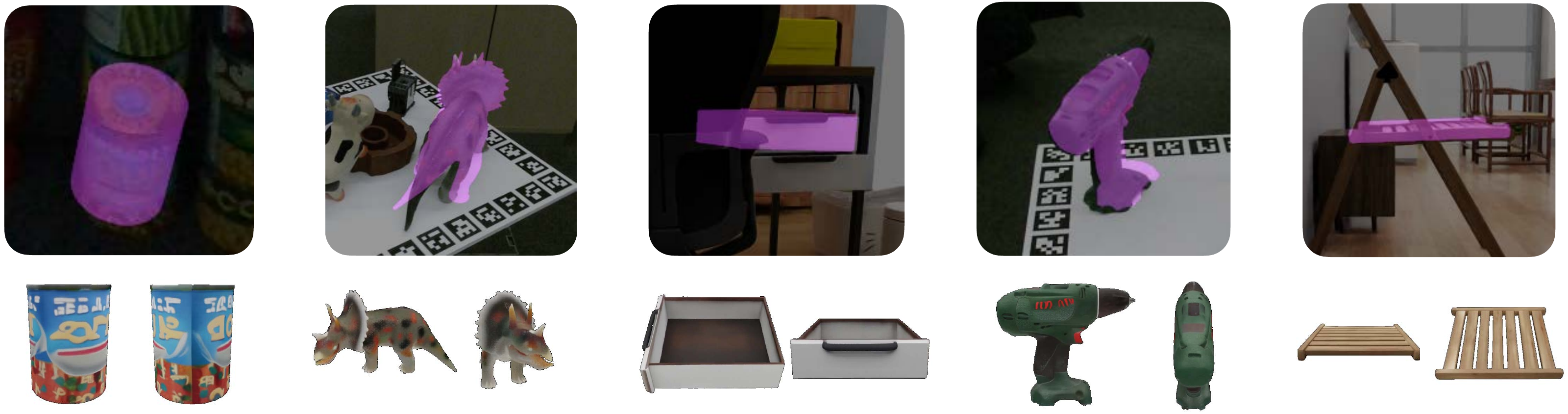} \\[2pt]
    \caption{\textbf{\method qualitative results}. We demonstrate that our method is robust to occlusions, handles symmetric objects, and generalizes to real-world data despite being trained exclusively on synthetic data.}
    \label{fig:qualitative}
\end{figure}

\myparagraph{\textbf{Pointmap conditioning.}}
Many generative reconstruction methods rely on post-hoc pose optimization since their models do not directly leverage depth information. To overcome this limitation, we introduce \emph{pointmap conditioning}, enabling the structure generator $\boldsymbol{\mathcal{G}}_{\mathrm{SP}}$ to utilize depth cues directly. The pointmap $\pointmap \in \mathbb{R}^{\imagedim \times \imagedim \times 3}$ is a convenient camera-invariant representation formed by recovering the missing spatial coordinates from the depth map $\depth \in \mathbb{R}^{\imagedim \times \imagedim}$ using camera intrinsics $\mathbf{K} \in \mathbb{R}^{3 \times 3}$. It is encoded through a learnable layer, and its feature map is added to the image features, providing explicit geometric grounding. This conditioning improves both pose accuracy and shape consistency without additional optimization. As depth can range drastically in the scene, we filter out all background pixels using provided object masks $\mask$: $\pointmap_{\text{obj}} = \mask \cdot \pointmap$. We further normalize the pointmap using its scale $s_{\text{obj}}$ and its translation $\boldsymbol{t}_{\text{obj}}$  to unify the input scale for our network. For translation, we use a robust estimate of the object center (median pointmap value on each dimension). For scale, we use the distance between the $5$-th and $95$-th percentile of point norms from the median center. The final pointmaps are obtained with $\pointmap^{\text{norm}}_{\text{obj}}
= \frac{\pointmap_{\text{obj}} - \boldsymbol{t}_{\text{obj}}}{s_{\text{obj}}}$, mapping the object into $[0,1]^3$. This way, background depth noise in the image does not affect the predictions, making the model more robust to real-world usage.~\looseness=-1

\subsubsection{High-Fidelity Asset Recovery.}
In the second stage, we generate the local latents $\{\boldsymbol{z}_i\}_{i=1}^{L}$ conditioned on the sparse structure and predicted pose using a sparse transformer $\boldsymbol{\mathcal{G}}_{\mathrm{L}}$. To enhance efficiency, we pack the latents within $2^3$ spatial neighborhoods using sparse convolutions~\cite{wang2017ocnn} before serialization, as in DiT~\cite{peebles2023scalable}. The packed sequence is processed through time-modulated transformer blocks, followed by a convolutional upsampling head with skip connections to preserve spatial detail. As in $\boldsymbol{\mathcal{G}}_{\mathrm{SP}}$, timesteps are integrated using AdaLN, and multimodal conditioning is applied via cross-attention layers. Critically, the predicted pose $\objectpose$ from Stage 1 is encoded through a learnable linear layer and concatenated with image, mask, and pointmap features. This pose conditioning is essential for symmetric objects with view-dependent appearance (e.g., cylindrical containers with labels), where only the pose provides the necessary grounding to generate $\boldsymbol{z}$ with appearance details consistent with the object's orientation. The resulting structured latents $\boldsymbol{z} = \{(\boldsymbol{z}_i, \boldsymbol{p}_i)\}_{i=1}^{L}$ are decoded by a mesh decoder $\boldsymbol{\mathcal{D}}_{\mathrm{M}}$, which extracts geometry via FlexiCubes~\cite{shen2023flexicubes}, and a Gaussian Splatting (GS) decoder $\boldsymbol{\mathcal{D}}_{\mathrm{GS}}$, which produces a set of colored 3D Gaussians capturing appearance. To obtain a textured mesh, the GS representation is rendered from multiple viewpoints and the resulting images are baked onto the mesh.

\myparagraph{\textbf{Training and losses.}}
Both $\boldsymbol{\mathcal{G}}_{\mathrm{SP}}$ and $\boldsymbol{\mathcal{G}}_{\mathrm{L}}$ are trained independently using the Conditional Flow Matching (CFM) objective from~\cite{lipman2023flow}. For $\boldsymbol{\mathcal{G}}_{\mathrm{SP}}$, we jointly optimize structure and pose using a weighted combination:
\begin{equation*}
    \mathcal{L}_{\mathrm{total}} = \mathcal{L}_{\mathrm{CFM}}(\boldsymbol{S}) + \alpha \cdot \mathcal{L}_{\mathrm{CFM}}(\tilde{\objectpose}),
\end{equation*}
where $\alpha = 0.01$ balances pose prediction with structure generation. We employ synthetic datasets with known ground-truth shapes and poses to supervise both spatial alignment and geometric reconstruction, ensuring consistency across the two generative stages.~\looseness=-1

\myparagraph{\textbf{Extension to Multiple Views.}}
 The vast majority of generative shape reconstruction methods resort to recovering shapes from a single image. This setup demonstrates exciting abilities of generative methods at recovering unobserved object parts using the learned object prior. However, practical real-world robotics and reconstruction setups commonly utilize multiple cameras allowing to alleviate uncertainty imposed by ambiguity in object symmetry and occlusion. To address this and increase the practical value of the method, we extend our training to the multi-view regime by adding an optional second image, pointmap, and mask tuple $\image^{(2)}$, $\pointmap^{(2)}$, $\mask^{(2)}$ with the corresponding pose $\objectpose^{(2)}$. The per-view conditioning features are concatenated along the sequence dimension, and a learnable frame token embedding is added to each view's patches so the cross-attention layers can distinguish their origin. Similarly, since the model now predicts one pose per view, each pose output token receives a learnable view id embedding to disambiguate the two predictions. During training, we drop the second view and its pose with probability $p_{\text{drop}}=0.33$, 
 allowing the network to leverage all available information while retaining single-view inference capability.

\subsection{RecGen Dataset.} Our RecGen dataset leverages 198K high-quality 3D assets from 6 public objects and parts datasets. In particular, we use objects assets from: Objaverse-XL~\cite{deitke2023objaversexl}, ABO~\cite{collins2022abo}, and HSSD~\cite{khanna2023hssd} and part assets from: PhysXNet~\cite{cao2025physx}, PartNext~\cite{wang2025partnext} and articulated parts from PartNet-Mobility~\cite{Xiang_2020_SAPIEN}. For the object-based datasets, we create compositional scenes where other assets from the same dataset were randomly placed in the scene to create natural occlusions and non-trivial depth patterns. For the Part-based scenes, we used a single object, as their parts are often severely self-occluded. Each scene is rendered into 20 images with random camera poses, resulting in a diverse set of viewpoints and lighting conditions. The dataset contains a total of 3.2M synthetically generated RGB images, depth maps, segmentation masks, GT poses, and stereo depth maps of 198K scenes with and without occlusions. For the training of the appearance generation, we excluded PartNet-Mobility and PhysXNet subsets due to the lower quality of the provided texture. Some sample assets from the 6 datasets are shown in \fig{fig:dataset_samples}.

\subsection{Implementation Details} \method adopts the rectified flow transformer architecture proposed in~\cite{xiang2025structured}. We use classifier-free guidance (CFG) with a drop rate of 0.1 and AdamW optimizer with a learning rate of 1e-4. The model is trained for 55K iterations with a batch size of 512 on 64 NVIDIA H100 GPUs. The training process takes approximately 48 hours. During inference, we use a CFG scale of 3.0 and perform 50 denoising steps.
All experiments are performed with the \texttt{TRELLIS-image-large}~\cite{xiang2025structured} model as the base representation,
starting from their pretrained network, which has around 1.2 billion parameters.

\section{Experiments}
\label{sec:experiments}

\begin{table}[t]
\centering
\scriptsize
\setlength{\tabcolsep}{12pt}
\renewcommand{\arraystretch}{1.15}
\caption{Quantitative comparison on object and part datasets, \method achieves the best performance on all metrics across all datasets.}
\resizebox{\columnwidth}{!}{%
\begin{tabular}{lllccccc}
\toprule
 & Dataset & Model & \makecell{$\text{CD}_{\text{norm}}$ \\ $(\downarrow)$} & \makecell{ADD-SB \\ $(\downarrow)$} & \makecell{ADD-SB \\ @0.1 $(\uparrow)$} & \makecell{ADD-SB \\ @0.05 $(\uparrow)$} & \makecell{DRE \\ @0.05 $(\uparrow)$} \\
\midrule
\multirow{18}{*}{\rotatebox{90}{Objects}} & \multirow{6}{*}{HB} & SceneComplete             & 0.234 & 0.258 & 65.2\% & 35.1\% & 0.0\% \\
 & & Any6D (InstantMesh)       & 0.074 & 0.111 & 68.6\% & 36.4\% & 33.6\% \\
 & & Any6D (Trellis)           & 0.106 & 0.157 & 47.8\% & 33.8\% & 26.8\% \\
 & & SAM3D                     & 0.033 & 0.062 & 92.4\% & 54.6\% & 34.6\% \\
 & & RecGen (1-view)           & \textbf{0.032} & \textbf{0.049} & \textbf{95.0\%} & \textbf{73.8\%} & \textbf{51.5\%} \\
\cmidrule(lr){3-8}
 & & \color{gray} RecGen (2-view)           & \color{gray} 0.029 & \color{gray} 0.048 & \color{gray} 95.4\% & \color{gray} 74.2\% & \color{gray} 50.9\% \\
\cmidrule(lr){2-8}
 & \multirow{6}{*}{ReOcS} & SceneComplete             & 0.764 & 0.774 & 38.2\% & 26.1\% & 0.0\% \\
 & & Any6D (InstantMesh)       & 0.055 & 0.066 & 89.5\% & 60.8\% & 60.5\% \\
 & & Any6D (Trellis)           & 0.068 & 0.088 & 75.5\% & 47.4\% & 47.1\% \\
 & & SAM3D                     & 0.026 & 0.057 & 96.2\% & 43.6\% & 25.8\% \\
 & & RecGen (1-view)           & \textbf{0.019} & \textbf{0.032} & \textbf{100.0\%} & \textbf{89.5\%} & \textbf{60.8\%} \\
\cmidrule(lr){3-8}
 & & \color{gray} RecGen (2-view)           & \color{gray} 0.018 & \color{gray} 0.032 & \color{gray} 99.7\% & \color{gray} 91.1\% & \color{gray} 62.4\% \\
\cmidrule(lr){2-8}
 & \multirow{6}{*}{LMO} & SceneComplete             & 0.186 & 0.222 & 50.0\% & 11.3\% & 0.0\% \\
 & & Any6D (InstantMesh)       & 0.100 & 0.148 & 42.2\% & 11.3\% & 19.0\% \\
 & & Any6D (Trellis)           & 0.116 & 0.196 & 29.6\% & 16.9\% & 15.5\% \\
 & & SAM3D                     & 0.057 & 0.110 & 64.1\% & 17.6\% & 34.5\% \\
 & & RecGen (1-view)           & \textbf{0.050} & \textbf{0.068} & \textbf{83.1\%} & \textbf{50.0\%} & \textbf{38.0\%} \\
\cmidrule(lr){3-8}
 & & \color{gray} RecGen (2-view)           & \color{gray} 0.056 & \color{gray} 0.075 & \color{gray} 83.1\% & \color{gray} 55.6\% & \color{gray} 37.3\% \\
\midrule
\multirow{6}{*}{\rotatebox{90}{Parts}} & \multirow{6}{*}{ArtVIP} & SceneComplete             & 0.189 & 0.201 & 57.2\% & 34.0\% & 0.6\% \\
 & & Any6D (InstantMesh)       & 0.089 & 0.100 & 61.3\% & 39.1\% & 16.2\% \\
 & & Any6D (Trellis)           & 0.090 & 0.106 & 58.0\% & 37.7\% & 16.8\% \\
 & & SAM3D                     & 0.056 & 0.073 & 79.2\% & 45.8\% & 22.6\% \\
 & & RecGen (1-view)           & \textbf{0.026} & \textbf{0.034} & \textbf{96.4\%} & \textbf{84.0\%} & \textbf{24.4\%} \\
\cmidrule(lr){3-8}
 & & \color{gray} RecGen (2-view)           & \color{gray} 0.024 & \color{gray} 0.032 & \color{gray} 96.4\% & \color{gray} 86.4\% & \color{gray} 24.8\% \\
\bottomrule
\end{tabular}
}
\label{tab:quant_results}
\end{table}

\myparagraph{\textbf{Evaluation datasets.}} We evaluate our method and baselines on four object-based datasets (LM-O~\cite{brachmann2014learning}, HB~\cite{kaskman2019homebreweddb}, HOPE~\cite{tyree2022hope}, and ReOcS~\cite{iwase2025zerograsp})  and one part-based dataset (ArtVIP~\cite{jin2025artvip}) for shape and pose estimation. The selected object datasets are widely used for benchmarking 6DoF pose estimation and provide GT meshes~\cite{hodan2018bop}. Each dataset represents distinct challenges: LM-O and HB include diverse objects and highly occluded scenes, while HOPE and ReOcS contain multiple symmetric objects with complex textures. In addition, these datasets are captured with different depth sensors (structured light, time of flight, and stereo) allowing us to evaluate the robustness of the baselines across sensor types. Detailed descriptions of each evaluation dataset are provided in \app{sec:object-based-eval}.

Since most of the real world pose estimation benchmarks are object-based and datasets with part annotations are scarce, we introduce a part-based benchmark derived from ArtVIP~\cite{jin2025artvip}, a collection of digital assets for high fidelity physical interaction in robot learning. The original dataset contains six static scenes; we extend it with six additional scenes featuring new objects. From these scenes, we select 284 object parts and render 924 high-quality RGB-D images. Further details on the benchmark construction are provided in \app{sec:part-based-eval}.

\myparagraph{\textbf{Evaluation metrics.}} 
To evaluate 6D pose estimation accuracy, we use the ADD-S metric. Since GT object meshes are generated rather than provided, we follow SAM3D~\cite{chen2025sam} and adopt a bidirectional variant of ADD-S (denoted \textit{ADD-SB}), which computes symmetric distances between the predicted and GT posed meshes. We report results at the standard $10\%$ object diameter threshold and additionally at $5\%$ ($ADD-SB@5\%$), as the former saturates on simpler datasets. 

To assess robustness to occlusions, we introduce the Diameter Relative Error (DRE) metric. Although estimating object size is straightforward for fully visible objects with depth, it becomes significantly more challenging under heavy partial occlusions, where only a small portion is observed. We define DRE as $e_d = |d_{\text{pred}} - d_{\text{gt}}| / d_{\text{gt}}$, where $d_{\text{pred}}$ and $d_{\text{gt}}$ denote the predicted and GT diameters. We report $DRE@0.05$, the fraction of samples with $e_d < 5\%$.

To evaluate surface reconstruction quality, we compute Chamfer Distance (CD) after ICP alignment to GT mesh and normalize it by the GT diameter to ensure equal weighting across object sizes. The ICP step helps disentangle shape errors from pose inaccuracies.

Finally, to assess the visual fidelity of the reconstructions, we render the predicted shapes from predefined views and report standard perceptual metrics PSNR, SSIM, and LPIPS.\looseness=-1

\myparagraph{\textbf{Baselines.}} 
We compare \method against baselines for model-free pose estimation~\cite{lee2025any6d}, scene completion~\cite{agarwal2024scenecomplete}, and 3D reconstruction from single images~\cite{chen2025sam}. 
Any6D~\cite{lee2025any6d} is a model-free pose estimation method that generates a mesh using InstantMesh~\cite{xu2024instantmesh} and refines the scale and pose via a full-to-partial matching strategy. Although the authors propose an anchor-query approach, we treat the two views as equivalent in our experiments. Additionally, we evaluate a variant using TRELLIS~\cite{xiang2025structured} for mesh generation.
For scene completion, we evaluate SceneComplete~\cite{agarwal2024scenecomplete}, which leverages inpainting to generate an occlusion-free mesh. The scale is estimated via feature matching, and the pose is refined using FoundationPose~\cite{wen2024foundationpose}.
Finally, we evaluate SAM3D~\cite{chen2025sam},  the closest related approach, as it simultaneously estimates both mesh and pose for objects. A detailed description of each baseline and its usage is provided in \App{sec:dataset-description}.\looseness=-1

\begin{figure}[t]
    \centering
    \includegraphics[width=\linewidth]{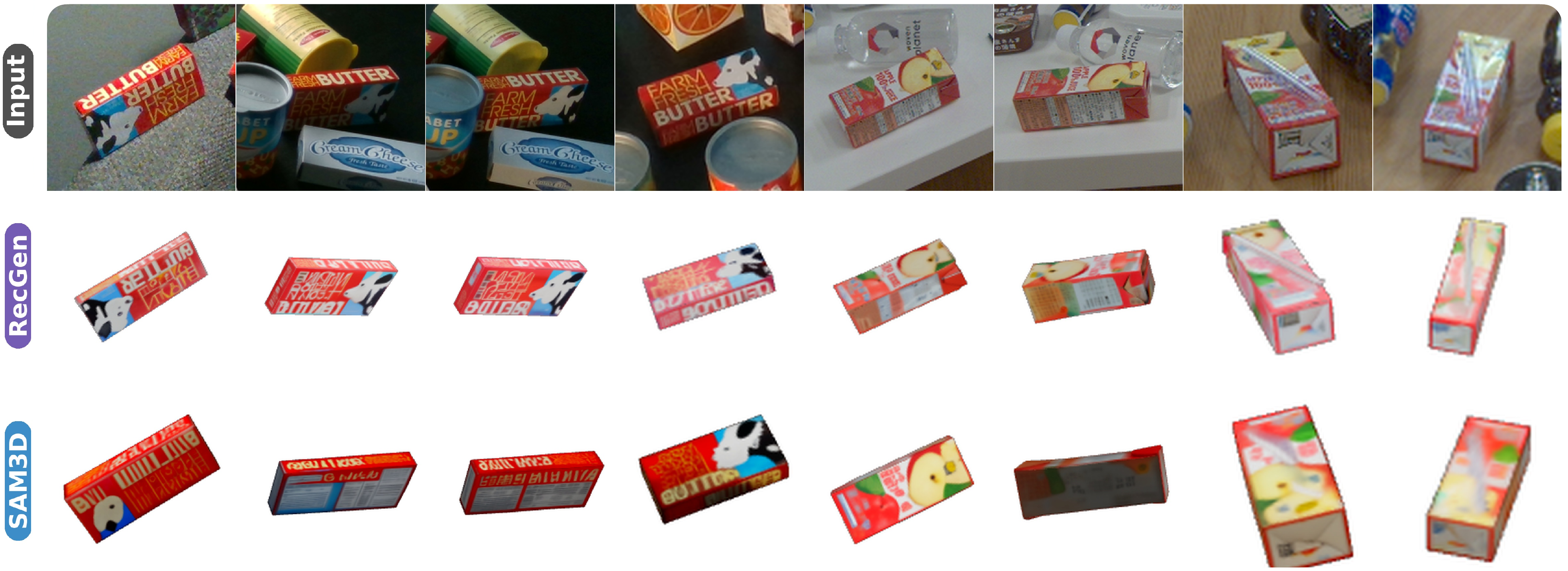}
    \caption{Qualitative comparison on symmetric objects. Our method generates textures consistent with the given pose, whereas SAM3D often produces incorrect textures because its appearance generation depends only on object shape, not pose.}
    \vspace{-5mm}
    \label{fig:symmetry_comparison}
\end{figure}

\subsection{Pose and Shape Estimation for Objects and Parts}
\Tab{tab:quant_results} reports shape quality ($\text{CD}_{\text{norm}}$) and pose estimation accuracy (ADD-SB) on both object-centric  and part-level benchmarks.

\begin{wrapfigure}{r}{0.38\linewidth}
    \centering
    \vspace{-1.2em}
    \includegraphics[width=\linewidth]{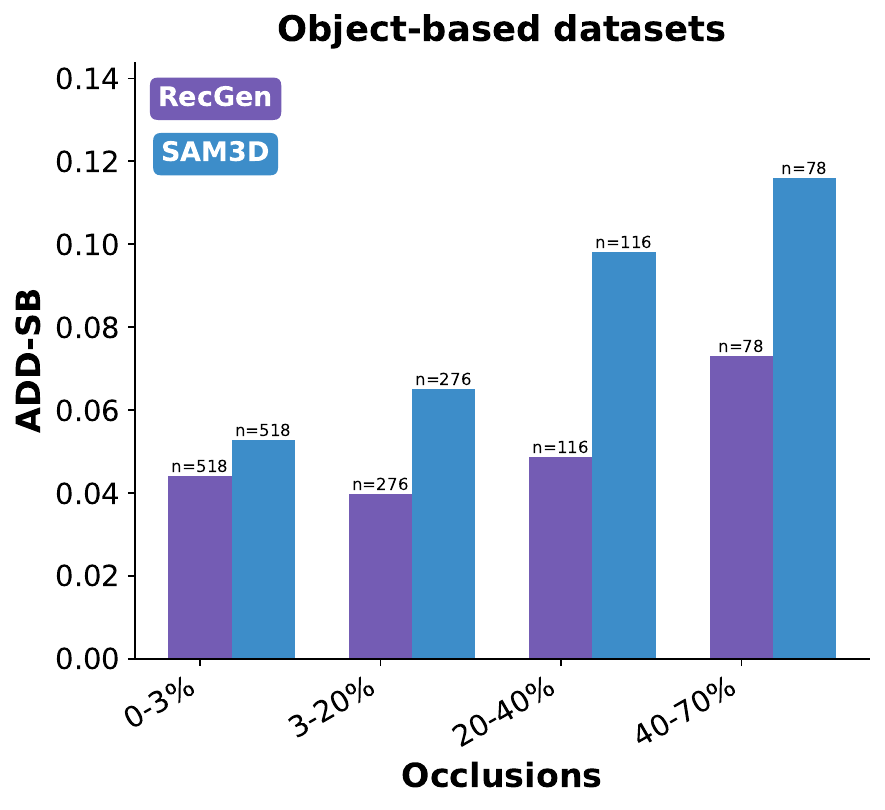}
    \caption{\textbf{Robustness to occlusions.} ADD-SB (lower is better) on object-based datasets (HB+LMO+ReOcS), binned by occlusion severity. \method's advantage widens as occlusion increases.}
    \label{fig:occlusion-main}
\end{wrapfigure}

Both \method and SAM3D outperform SceneComplete and Any6D, highlighting the importance of joint shape and pose training on large-scale datasets with occlusion.
On object-centric benchmarks, \method~(1-view) achieves an average $\text{CD}_{\text{norm}}$ of $0.033$ vs.\ $0.039$ for SAM3D and $0.076$ for Any6D.
The 2-view variant further improves shape generation ($\text{CD}_{\text{norm}}$ is $0.034$ for objects, $0.024$ for parts), enabling more accurate reconstruction in standard robotics setups where more than one RGB-D camera is available~\cite{khazatsky2024droid}. Inference-time pose-selection and a multi-sample selection strategies that further improve the two-view results are discussed in \app{sec:multiview-pose-selection} and \app{sec:multisample-selection}.\looseness=-1

With respect to  pose estimation, \method outperforms all baselines, including the state-of-the-art SAM3D. On object-centric benchmarks, \method~(1-view) reaches $92.7\%$ ADD-SB @$0.1$ and $71.1\%$ @$0.05$ on average, compared to an average of $84.2\%$ / $38.6\%$ for SAM3D---nearly doubling accuracy at the stricter threshold. The 2-view variant enhances the average performance to $73.6\%$ @$0.05$ (see also \tab{tab:multiview_pose_selection}). \method's ability to accurately predict full object scale, even on occluded samples, can be further seen in the DRE metric with significant improvements over SAM3D, as qualitatively visible in \fig{fig:qualitative}. Robustness to occlusion severity is highlighted in \fig{fig:occlusion-main}: on object-based datasets the gap to SAM3D widens from $0.044$ vs.\ $0.053$ at 0--3\% occlusion to $0.073$ vs.\ $0.116$ at 40--70\% occlusion (a 37\% relative improvement). A more complete analysis including Chamfer distance and the parts-based AV dataset is provided in \app{sec:occlusion-analysis}; a per-object breakdown on HB is reported in \app{sec:per-object-hb}, and additional qualitative comparisons are shown in \app{sec:object-based-appendix}.\looseness=-1

Reconstructing and localizing articulated object parts is a particularly challenging task that requires fine-grained geometric understanding. Although part geometries are often simpler than full objects, their recovery requires inferring the underlying shape under severe self-occlusion and benefits greatly from part-aware joint shape and pose prediction training.
\method outperforms all baselines by a large margin on the ArtVIP part-level benchmark: for part-shape reconstruction, \method~(1-view) reduces $\text{CD}_{\text{norm}}$ by half compared to SAM3D ($0.056 \rightarrow 0.026$); for part-pose estimation, it improves ADD-SB @$0.05$ by $+38.2$pp ($45.8\% \rightarrow 84.0\%$), establishing state-of-the-art in both tasks. This capability makes \method ideal for extending Real-to-Sim-to-Real~\cite{barcellona2025dream} beyond object rearrangement to articulated object manipulation. Additional qualitative comparisons on ArtVIP parts are shown in \App{sec:part-based-appendix}.

\begin{table}[t]
\centering
\scriptsize
\setlength{\tabcolsep}{12pt}
\renewcommand{\arraystretch}{1.15}

\caption{Perception quality comparison before and after ICP.} %
\resizebox{\columnwidth}{!}{%
\begin{tabular}{llcccccc}
\toprule
Dataset & Model & \multicolumn{3}{c}{Before ICP} & \multicolumn{3}{c}{After ICP} \\
\cmidrule(lr){3-5} \cmidrule(lr){6-8}
 &  & \makecell{LPIPS $(\downarrow)$} & \makecell{SSIM $(\uparrow)$} & \makecell{PSNR $(\uparrow)$} & \makecell{LPIPS $(\downarrow)$} & \makecell{SSIM $(\uparrow)$} & \makecell{PSNR $(\uparrow)$} \\
\midrule
\multirow{5}{*}{LMO+HB+HOPE} & Any6D (InstantMesh)       & 0.225 & \textbf{0.835} & 15.46 & 0.230 & 0.825 & 15.20 \\
 & Any6D (Trellis)           & 0.263 & 0.829 & 14.56 & 0.257 & 0.820 & 14.48 \\
 & SAM3D                     & 0.219 & 0.821 & 15.72 & \textbf{0.161} & \textbf{0.841} & \textbf{17.42} \\
 & RecGen (1-view)           & \textbf{0.199} & 0.825 & \textbf{15.85} & 0.170 & 0.834 & 16.54 \\
\cmidrule(lr){2-8}
 & \color{gray} RecGen (2-view)           & \color{gray} 0.199 & \color{gray} 0.824 & \color{gray} 15.82 & \color{gray} 0.166 & \color{gray} 0.835 & \color{gray} 16.62 \\
\midrule
\multirow{5}{*}{Symmetric} & Any6D (InstantMesh)       & 0.193 & \textbf{0.834} & 16.31 & 0.201 & 0.822 & 15.96 \\
 & Any6D (Trellis)           & 0.190 & 0.834 & \textbf{16.45} & 0.187 & \textbf{0.829} & 16.50 \\
 & SAM3D                     & 0.201 & 0.815 & 16.02 & 0.156 & 0.828 & \textbf{17.21} \\
 & RecGen (1-view)           & \textbf{0.170} & 0.816 & 15.63 & \textbf{0.142} & 0.827 & 16.12 \\
\cmidrule(lr){2-8}
 & \color{gray} RecGen (2-view)           & \color{gray} 0.172 & \color{gray} 0.817 & \color{gray} 15.59 & \color{gray} 0.144 & \color{gray} 0.830 & \color{gray} 16.08 \\
\bottomrule
\end{tabular}
}
\label{tab:perception_results_v2}
\end{table}

\subsection{Posed Appearance Generation}

To evaluate the quality of our pose-aware appearance generation, we use the HB, LM-O, and HOPE datasets, which provide textures for the GT object meshes. For each predicted sample, we transform it using the predicted pose and render the generated appearance, then compare it with the GT object transformed with the GT pose and rendered from the same view. To disentangle the effects of pose estimation errors, we  perform ICP alignment using GT shapes as references.\looseness=-1

\begin{wrapfigure}{r}{0.27\linewidth}
    \centering
    \includegraphics[width=\linewidth]{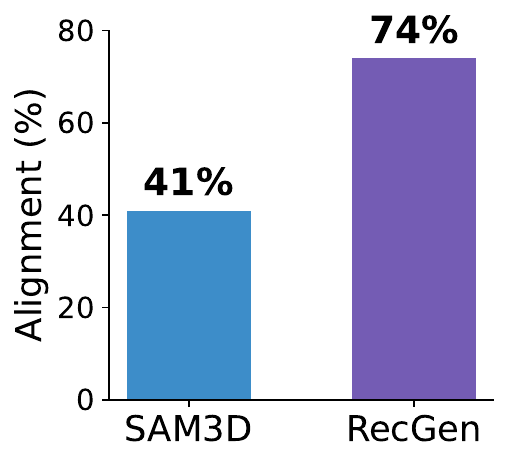}
    \caption{VLM-based evaluation of texture orientation alignment to the GT mesh on symmetric objects.}
    \label{fig:vlm_orientation}
    \vspace{-1.5em}
\end{wrapfigure}

In \tab{tab:perception_results_v2}, we observe that before additional ICP alignment \method outperforms the other baselines on average, demonstrating how the combined pose, shape, and appearance estimation leads to a much more faithful overall scene reconstruction. After ICP refinement, both \method and SAM3D perform significantly better than other baselines and perform comparably to each other.

We expect that a larger training dataset and additional usage of the depth during training of the encoder-decoder (Depth-VAE 
from SAM3D), as well as integrating multi-resolution training (TRELLIS 2), can further improve the quality of \method's pose-aware and multi-view appearance generation.

To additionally study the usage of the poses during appearance generation, we evaluated all methods on a subset of symmetric 
objects from HOPE and HB. We see a relative improvement in the perceptual similarity measured by the LPIPS metric (see \tab{tab:perception_results_v2}). To verify the source of the improvement, we additionally perform a VLM-based classification of the orientation alignment based on two images using the GPT-5 model, comparing the GT-posed and rendered objects with the ICP-aligned prediction from the model.  As depicted in \fig{fig:vlm_orientation}, \method surpasses SAM3D in object texture orientation alignment by a large margin. We attribute this improvement to our pose-conditioned formulation, which enables more accurate texture recovery for symmetric objects by properly aligning textures with input views. Some qualitative results are shown in \fig{fig:symmetry_comparison}; a per-object breakdown of the VLM orientation evaluation and additional qualitative examples are provided in \app{sec:symmetric-objects-appendix}.\looseness=-1

\subsection{Ablation study}

\begin{table}[t]
\centering
\small
\setlength{\tabcolsep}{24pt}
\renewcommand{\arraystretch}{1.15}

\caption{\textbf{Ablation study for joint shape and pose generation.} Object-level results are reported on HB, LM-O, and ReOcS, and part-level results on ArtVIP. Values are shown as mean / median. Differences from the full model are highlighted in {\color{ForestGreen!100} green (better)}, {\color{BrickRed!100} red (worse)}, and {\color{gray} gray (similar)}.}
\resizebox{\columnwidth}{!}{%
\begin{tabular}{lcccc}
\toprule
Variant & \multicolumn{2}{c}{Objects-centric} & \multicolumn{2}{c}{Part-centric} \\
\cmidrule(lr){2-3} \cmidrule(lr){4-5}
 & $\text{CD}_{\text{norm}}$ $(\downarrow)$ & ADD-SB $(\downarrow)$ & $\text{CD}_{\text{norm}}$ $(\downarrow)$ & ADD-SB $(\downarrow)$ \\
\midrule
Full model & 0.042\,/\,0.023 & 0.062\,/\,0.037 & 0.033\,/\,0.020 & 0.043\,/\,0.028 \\
w/o stereo & \color{BrickRed!100} 0.048\,/\,0.030 & \color{BrickRed!100} 0.078\,/\,0.050 & \color{ForestGreen!100} 0.030\,/\,0.018 & \color{ForestGreen!100} 0.039\,/\,0.027 \\
w/o norm & \color{gray} 0.042\,/\,0.026 & \color{BrickRed!100} 0.074\,/\,0.048 & \color{BrickRed!100} 0.038\,/\,0.025 & \color{BrickRed!100} 0.056\,/\,0.041 \\
w/o part-centric & \color{gray} 0.040\,/\,0.023 & \color{gray} 0.060\,/\,0.037 & \color{BrickRed!100} 0.073\,/\,0.037 & \color{BrickRed!100} 0.086\,/\,0.048 \\
w/o pretraining & \color{BrickRed!100} 0.044\,/\,0.031 & \color{BrickRed!100} 0.067\,/\,0.046 & \color{BrickRed!100} 0.044\,/\,0.028 & \color{BrickRed!100} 0.056\,/\,0.036 \\
\bottomrule
\end{tabular}
}
\label{tab:ablation_ss_agg}
\end{table}

The ablation study in \tab{tab:ablation_ss_agg} validates our design choices, showing that the full model achieves robustness to real-world conditions and generality across object- and part-level reasoning. Given the computational constraints, we train a base and ablated \method models for 150K iterations with a batch size of 64.\looseness=-1 
 
\myparagraph{Stereo Noise.} Removal of stereo noise augmentation substantially degrades object-centric metrics (CD $0.048$ vs. $0.042$, ADD-SB $0.078$ vs. 0.062), as the model becomes less resilient to real-world sensor noise. The effect is negligible on synthetic ArtVIP data, where depth is noise-free. Disabling pose normalization has little impact on shape quality (CD remains at 0.042), but significantly hurts pose estimation across both settings (ADD-SB rises to 0.074 and 0.056 from 0.062 and 0.043, respectively), confirming that normalization simplifies the pose learning task and makes joint optimization more stable. Together, stereo augmentation and pose normalization provide complementary robustness — the former at the input level and the latter at the representation level.

\myparagraph{Generality.} Training without part-level data matches the full model on object-centric benchmarks (CD $0.040$, ADD-SB $0.060$) but degrades substantially on ArtVIP (CD $0.073$ vs. $0.033$, ADD-SB $0.086$ vs. $0.043$). Part supervision is necessary for articulated structures while not compromising object-level performance.

\myparagraph{Pretraining} and thereafter initializing weights   improves all metrics, providing a strong geometric prior for shape reconstruction and joint pose estimation.

Finally, beyond accuracy, we find that \method is also substantially more efficient than SAM3D at inference time ($1.8\times$ faster, $1.6\times$ less GPU memory); a detailed efficiency comparison is reported in \App{sec:efficiency}. Limitations and future directions are discussed in \App{sec:limitations}.

\section{Conclusion}
\label{sec:conclusion}
In this paper, we propose \method, a generalist scene completion framework that recovers entirely multi-object shapes from partial input observations. \method addresses several long-standing challenges in computer vision: it is robust to occlusions, handles symmetric objects and relative object-parts, generalizes to real-world data despite being trained solely on synthetic data and works robustly across different real-world RGB-D sensors. Unlike many competing approaches, \method extends to multiple views, making it more suitable for real-world applications. We demonstrate that \method outperforms the competitive SAM3D by 30.1\% in geometric shape quality, 9.1\% in texture reconstruction, and 33.9\% in pose estimation, while using 80\% less training data meshes. We believe that \method can serve as an easy-to-deploy and easy-to-build-on framework for advancing real-to-sim reconstruction in robotics and other fields.

\newpage
\section*{Acknowledgements}
\label{sec:acknowledgements}
Andrii Zadaianchuk is funded by the
European Union (ERC, EVA, 950086). 

\section*{Contributions}
\label{sec:contributions}
Andrii Zadaianchuk was the main contributor and was responsible for conceptualization, methodology, training data generation, code development, model training, evaluations development, writing, and project direction. Sergey Zakharov provided main supervision and contributed to conceptualization, methodology, training data generation, code development, model training, and project direction. Leonardo Barcellona was a core contributor and was involved in conceptualization, evaluations development, part-based evaluation dataset, evaluation of the baselines, and project direction. Christian Gumbsch contributed to validation and formal analysis and provided important feedback during the project. Lennard Schuenemann contributed to validation and formal analysis. Zehao Wang contributed to visualization and writing. Muhammad Zubair Irshad, Fabien Despinoy, Rahaf Aljundi, and Stratis Gavves contributed to writing, review, editing and provided important feedback during the project.

\printbibliography

\newpage
\renewcommand{\thetable}{S\arabic{table}}
\renewcommand{\thefigure}{S\arabic{figure}}
\renewcommand{\theequation}{S\arabic{equation}}
\setcounter{table}{0}
\setcounter{figure}{0}
\setcounter{equation}{0}
\appendix

\maketitlesupplementary

In these supplementary materials, we present an additional detailed analysis of the \method performance. We show that:

\begin{enumerate}
    \item \method is significantly more robust to partial and severe occlusions than prior work, with performance gaps widening as occlusion increases~(\app{sec:occlusion-analysis}).
    \item The performance gains are consistent across individual object instances, as demonstrated by a per-object breakdown on HB~(\app{sec:per-object-hb}).
    \item The pose-conditioned appearance generation of \method resolves symmetry ambiguities more reliably than pose-agnostic baselines~(\app{sec:symmetric-objects-appendix}).
    \item Simple inference-time strategies, such as multi-view pose selection and multi-sample alignment-based selection, further improve reconstruction quality without retraining~(\app{sec:multiview-pose-selection} and \app{sec:multisample-selection}).
    \item \method achieves superior computational efficiency compared to SAM3D, requiring less memory and runtime while maintaining higher accuracy~(\app{sec:efficiency}).
\end{enumerate}

In addition, in \sect{sec:datasets-appendix}, we provide additional details about the training dataset construction, the proposed evaluation benchmark, and a comprehensive description of the baseline methods and their usage.

Finally, we discuss \method limitations and future work and show additional qualitative results on both object-based (\sect{sec:object-based-appendix}) and part-based datasets (\sect{sec:part-based-appendix}).

\section{Additional Analysis of the \method Performance}
\subsection{Robustness to Occlusions}
\label{sec:occlusion-analysis}

We analyze how pose and shape estimation degrade as the target object becomes increasingly occluded in the input view.
For each test sample, we compute the \emph{occlusion fraction} -- the ratio of occluded object pixels to total object pixels in the input image, using the ground-truth segmentation masks provided by each dataset. We partition samples into four occlusion bins: 0--3\% (nearly fully visible), 3--20\%, 20--40\%, and 40--70\% (severely occluded). We then report the mean ADD-SB and normalized Chamfer scores for \method and the best performing baseline, SAM3D, within each bin.
Results are aggregated into two groups: \emph{object-based} datasets (HB, LMO, and ReOcS; 994 samples) and the \emph{parts-based} dataset (ArtVIP; 500 samples).

\begin{figure*}[!t]
    \centering
    \begin{subfigure}[b]{\textwidth}
        \centering
        \includegraphics[width=\textwidth]{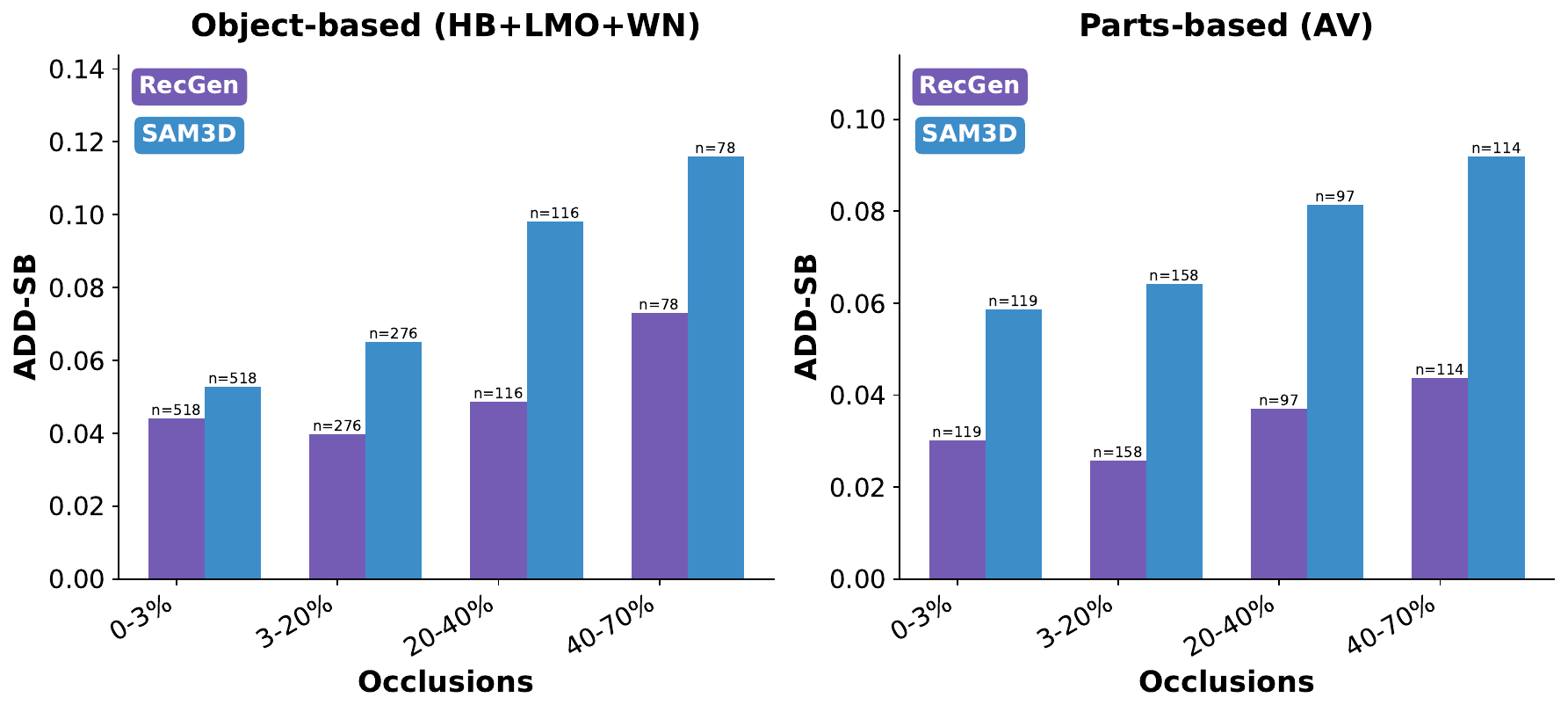}
        \caption{ADD-SB (lower is better) by occlusion severity.}
        \label{fig:occlusion-addss-grouped}
    \end{subfigure}
    \vspace{0.3em}
    \begin{subfigure}[b]{\textwidth}
        \centering
        \includegraphics[width=\textwidth]{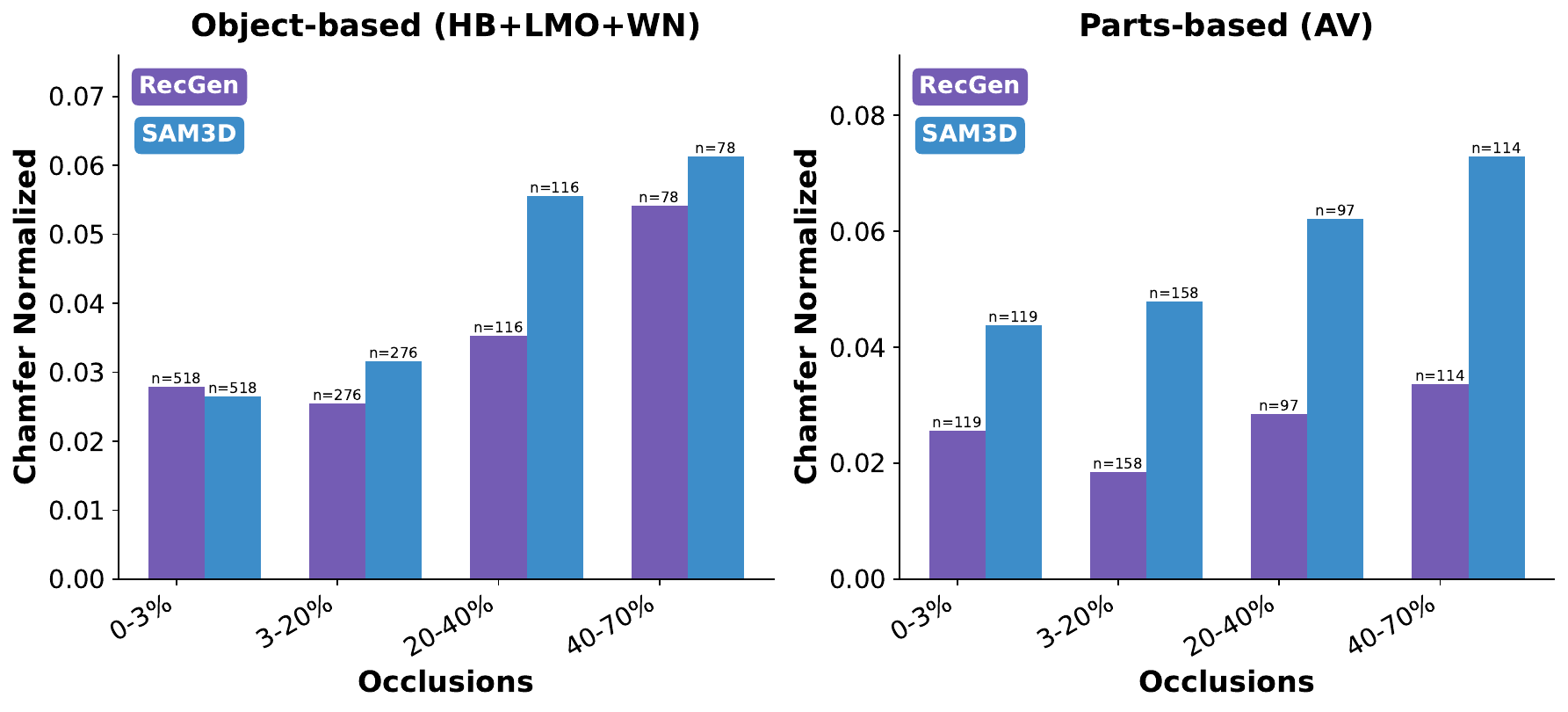}
        \caption{Normalized Chamfer distance (lower is better) by occlusion severity.}
        \label{fig:occlusion-chamfer-grouped}
    \end{subfigure}
    \caption{\textbf{Reconstruction quality vs.\ occlusion severity.} Samples are binned by the fraction of the target object visible in the input image. RecGen degrades gracefully as occlusion increases, while SAM3D's error grows substantially. Left: object-based datasets (HB+LMO+ReOcS). Right: parts-based dataset (AV).}
    \label{fig:occlusion-analysis}
\end{figure*}

\Fig{fig:occlusion-analysis} shows reconstruction quality as a function of occlusion severity.
On object-based datasets (\fig{fig:occlusion-addss-grouped}, left), \method consistently outperforms SAM3D in ADD-SB across all occlusion levels, and the gap widens as occlusion increases: at 0--3\% occlusion the difference is modest ($0.044$ vs.\ $0.053$), but at 40--70\% occlusion \method achieves $0.073$ compared to $0.116$ for SAM3D---a 37\% relative improvement.
The normalized Chamfer distance (\fig{fig:occlusion-chamfer-grouped}) tells a similar story, with both methods performing comparably on fully visible objects but \method maintaining lower error under heavy occlusion.

On the parts-based AV dataset (\fig{fig:occlusion-addss-grouped}, right), the advantage of \method is even more pronounced: \method's ADD-SB degrades only mildly from 0--3\% to 40--70\% occlusion, changing by $0.014$ (from $0.030$ to $0.044$), while SAM3D's error increases more drastically by $0.033$ (from $0.059$ to $0.092$). The same trend holds for Chamfer, where \method outperforms SAM3D by roughly $2\times$ across all occlusion bins.

These results suggest that \method's generative pose estimation combined with robust and object-centric scene normalization as well as usage of the real-world depth sensors during training are highly effective for improving real-world pose and shape estimation though occlusions.

\subsection{Per-Object Analysis on HB}
\label{sec:per-object-hb}

We present a detailed per-object breakdown of shape and pose metrics on the HB dataset, comparing \method to SAM3D across all 33 objects.
We report Chamfer Distance (normalized by object diameter) as a shape quality metric and ADD-SB as a pose accuracy metric; both are lower-is-better. Three outlier samples (2 of SAM3D for \texttt{object 21} and 1 of \method for \texttt{object 16}) with ADD-SB~$\geq 0.6$ are excluded from both methods symmetrically.

\begin{figure*}[!t]
    \centering
    \includegraphics[width=\textwidth]{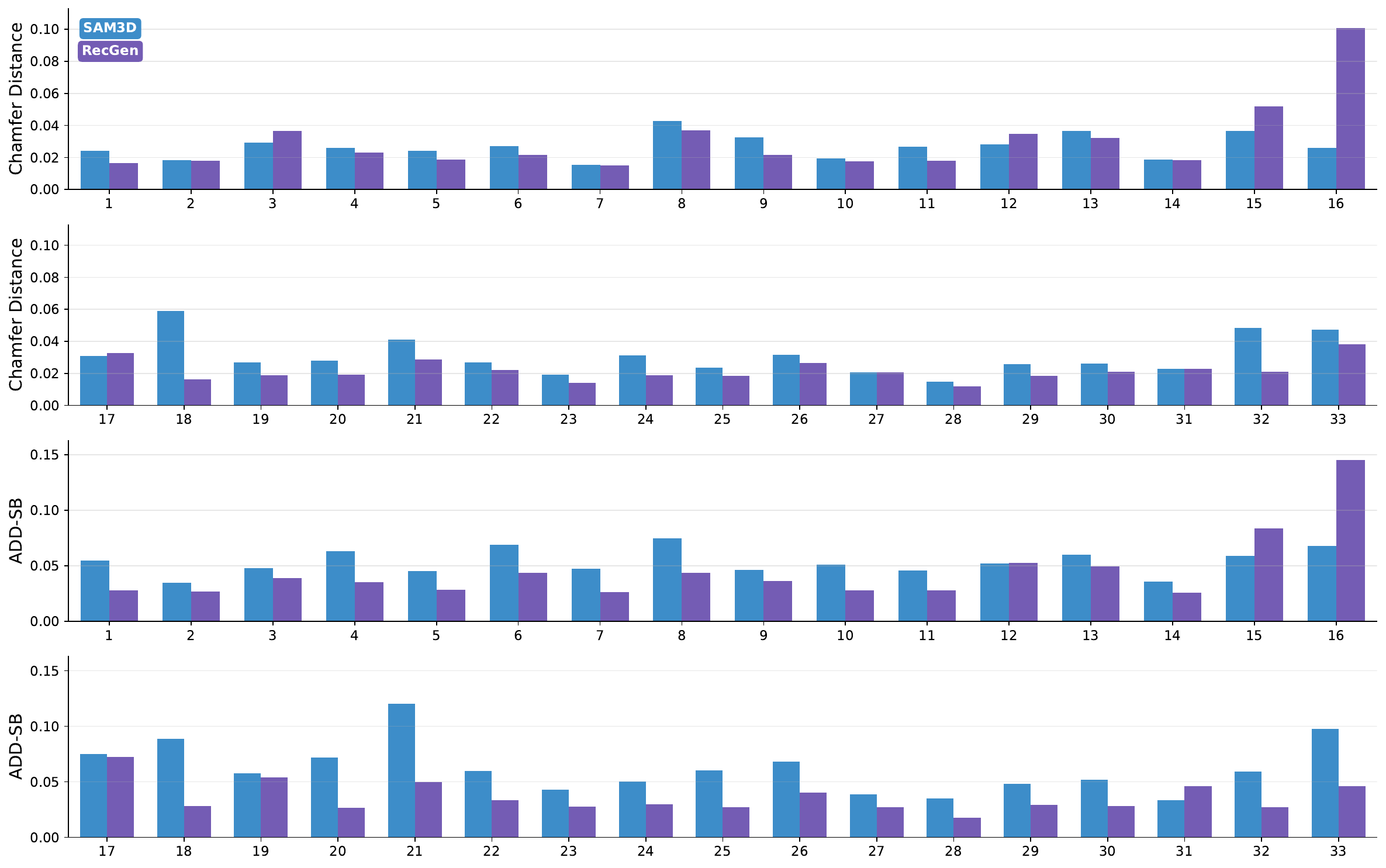}
    \caption{\textbf{Per-object comparison on HB dataset.} Chamfer Distance (top two rows) and ADD-SB (bottom two rows) for each of the 33 HB objects. Lower is better. \method (purple) outperforms SAM3D (blue) on 26/33 objects for shape and 29/33 for pose, while both have one outlier object on which they perform significantly worse than on other objects.}
    \label{fig:per-object-hb}
\end{figure*}

\Fig{fig:per-object-hb} visualizes the per-object comparison.
\method achieves lower ADD-SB on \textbf{29 out of 33} objects and lower Chamfer distance on \textbf{26 out of 33} objects, demonstrating consistent improvement across the majority of object instances. The four objects where SAM3D achieves better ADD-SB (objects 12, 15, 16, 31) tend to be cases where \method occasionally produces shape artifacts that affect pose alignment, while the underlying geometry predicted by SAM3D happens to be more stable for these specific instances. Notably, objects 15 and 16 are the only cases where SAM3D outperforms \method on \emph{both} metrics simultaneously.

\subsection{Symmetric Objects}
\label{sec:symmetric-objects-appendix}

Geometrically symmetric objects pose a unique challenge for joint shape and appearance reconstruction: because the object geometry is identical under some rotations, the predicted mesh can appear correct in terms of shape yet have its texture placed on the wrong side.
Methods that generate appearance independently of pose, such as SAM3D~\cite{chen2025sam}, are particularly susceptible to this failure mode, as they have no mechanism to resolve which face of the object is visible in the input view.
\method addresses this through its pose-conditioned formulation, which conditions appearance generation on the estimated pose, enabling the model to assign textures consistently with the observed viewpoint.

\begin{figure}[!t]
    \centering
    \includegraphics[width=0.85\linewidth]{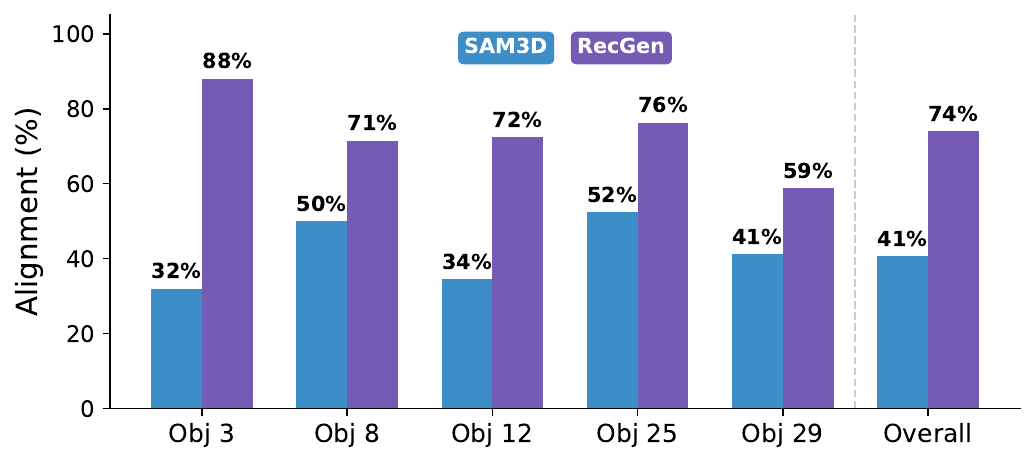}
    \caption{\textbf{Per-object VLM orientation alignment.} Alignment rates for each symmetric object, evaluated by GPT-5. \method consistently outperforms SAM3D across all objects. Overall alignment of \method on 5 objects~(106 input images) is 74\% vs.\ SAM3D 41\%.}
    \label{fig:vlm_per_object}
\end{figure}

To quantify how well the appearance generation network is using pose information, we perform a VLM-based orientation evaluation using GPT-5: for each symmetric object sample, we render the GT-posed object and the posed and additionally ICP-aligned prediction side by side and query the model whether the dominant visual regions (color blocks, labels, graphics) occupy the same spatial positions in both images.
\Fig{fig:vlm_per_object} breaks down the per-object alignment rates across five symmetric objects from HOPE (objects 3, 8, 12, 25) and HB (object 29).
\method outperforms SAM3D on every evaluated object, with the largest margin on HOPE object~3 (88\% vs.\ 32\%), where the texture is asymmetric along every axis, resulting in many plausible but incorrect orientations for pose-agnostic methods.
The overall alignment rate is 74\% for \method vs.\ 41\% for SAM3D.

\begin{figure*}[!t]
    \centering
    \includegraphics[width=\textwidth]{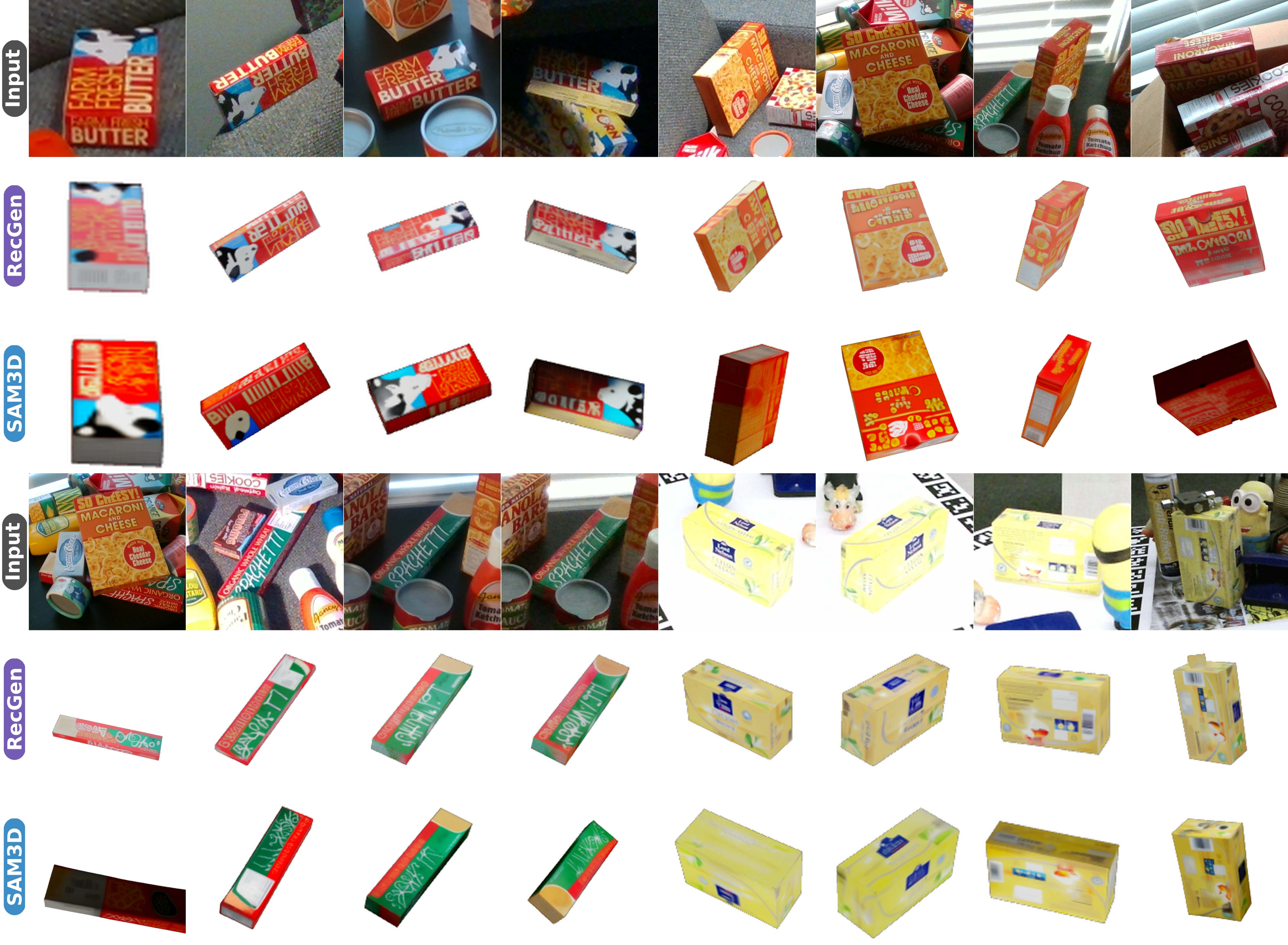}
    \caption{\textbf{Appearance generation for objects with symmetric shapes.} Top block: HOPE objects 3 and 12). Bottom block: HOPE object 25 and HB object 29. For each block: input image (top), \method reconstruction (middle), SAM3D reconstruction (bottom). \method produces textures consistent with the ground-truth orientation across diverse symmetric objects.}
    \label{fig:symmetry_comparison_appendix}
\end{figure*}

\Fig{fig:symmetry_comparison_appendix} provides additional qualitative examples spanning three HOPE objects (3, 12, 25) and HB object~29.
\method consistently reconstructs textures aligned with the ground-truth orientation, while SAM3D frequently produces flipped or misaligned textures.

\section{Inference Optimizations}

\subsection{Multi-view Pose Selection}
\label{sec:multiview-pose-selection}

\begin{table}[t]
\centering
\small
\setlength{\tabcolsep}{4pt}
\renewcommand{\arraystretch}{1.15}
\caption{Effect of multi-view pose selection. When two views are available, \method predicts two candidate poses. \emph{Single-view alignment} scores each pose against its own view's pointmap in metric camera space. \emph{Cross-view alignment} additionally uses GT relative camera poses to score each candidate against both views' pointmaps. Oracle selects the pose with lowest GT Chamfer distance.}
\resizebox{\columnwidth}{!}{%
\begin{tabular}{lcccccccccc}
\toprule
 & \multicolumn{2}{c}{HB} & \multicolumn{2}{c}{LMO} & \multicolumn{2}{c}{ReOcS} & \multicolumn{2}{c}{ArtVIP} & \multicolumn{2}{c}{Avg.} \\
\cmidrule(lr){2-3} \cmidrule(lr){4-5} \cmidrule(lr){6-7} \cmidrule(lr){8-9} \cmidrule(lr){10-11}
Method & $\text{CD}_n$ & ADD-SB & $\text{CD}_n$ & ADD-SB & $\text{CD}_n$ & ADD-SB & $\text{CD}_n$ & ADD-SB & $\text{CD}_n$ & ADD-SB \\
\midrule
\method~(1-view) & 0.032 & 0.049 & 0.051 & 0.068 & 0.019 & \textbf{0.032} & 0.026 & 0.034 & 0.032 & 0.046 \\
\method~(2-view) & 0.029 & 0.048 & 0.056 & 0.075 & \textbf{0.018} & \textbf{0.032} & 0.024 & 0.032 & 0.032 & 0.047 \\
\method~(2-view, single-view) & \textbf{0.026} & 0.043 & \textbf{0.043} & 0.059 & 0.019 & 0.033 & 0.023 & 0.030 & 0.028 & 0.041 \\
\method~(2-view, cross-view) & \textbf{0.026} & \textbf{0.042} & \textbf{0.043} & \textbf{0.057} & \textbf{0.018} & \textbf{0.032} & \textbf{0.022} & \textbf{0.029} & \textbf{0.027} & \textbf{0.040} \\
\cmidrule(lr){1-11}
\color{gray} \method~(2-view, oracle) & \color{gray} 0.024 & \color{gray} 0.039 & \color{gray} 0.039 & \color{gray} 0.054 & \color{gray} 0.017 & \color{gray} 0.030 & \color{gray} 0.021 & \color{gray} 0.028 & \color{gray} 0.025 & \color{gray} 0.038 \\
\bottomrule
\end{tabular}
}
\label{tab:multiview_pose_selection}
\end{table}

In the two-view setting, \method predicts one 6DoF pose per input view, yielding two candidate poses $\objectpose^{(1)}$ and $\objectpose^{(2)}$ for the same reconstructed mesh $\shape$. In the main paper, we report results using only the first pose $\objectpose^{(1)}$. Here, we investigate whether an automatic selection strategy can consistently pick the better candidate at inference time, without access to GT meshes.

We propose a pointmap-based pose selection strategy that operates in metric camera space. For each candidate pose $\objectpose^{(k)}$, we transform the predicted mesh into the camera coordinate frame of view~$k$ using the inverse of the pointmap normalization transform, then compute one-directional nearest-neighbor distances from each pointmap point to the mesh surface. A trimmed mean (removing the top 10\% of distances) provides robustness to partial visibility. We consider two variants:
\begin{itemize}[nosep,leftmargin=*]
    \item \emph{Single-view alignment}: each pose $\objectpose^{(k)}$ is scored solely by its alignment to its own view's pointmap. The pose with the lower score is selected.
    \item \emph{Cross-view alignment}: given the GT relative camera pose $\objectpose_{\text{rel}} = \objectpose^{(i)}_{\text{cam}} \circ (\objectpose^{(j)}_{\text{cam}})^{-1}$, each candidate is additionally scored against the other view's pointmap by transforming the mesh into the other camera frame. The per-view scores are averaged and the pose with the lower combined score is selected.
\end{itemize}

\Tab{tab:multiview_pose_selection} reports shape quality ($\text{CD}_n$) and pose accuracy (ADD-SB) for five configurations: single-view, two-view with the first pose, single-view alignment, cross-view alignment, and the oracle (GT-best).
Both selection strategies substantially improve over the first-pose baseline on three of four datasets, with the largest gains on LMO ($\text{CD}_n$: $0.056 \rightarrow 0.043$, a 23\% reduction) and HB ($\text{CD}_n$: $0.029 \rightarrow 0.026$, 10\% reduction). Cross-view alignment, which leverages GT relative camera poses, achieves the best average performance (avg.\ $\text{CD}_n$: $0.027$, ADD-SB: $0.040$), outperforming single-view alignment (avg.\ $\text{CD}_n$: $0.028$, ADD-SB: $0.041$); both clearly improve over the first-pose baseline (avg.\ $\text{CD}_n$: $0.032$, ADD-SB: $0.047$). Notably, on LMO, where the two-view first-pose result is worse than single-view ($0.056$ vs.\ $0.051$), both selection strategies recover and surpass it, demonstrating that the second pose provides complementary information that the selection mechanism can exploit. On ReOcS, where there are almost no occlusions, using the first pose is comparable or better than selecting between poses, as the original views already provide sufficient information for accurate pose prediction. Overall, we recommend using such pose selection in cases where severe occlusions are possible, as there it could be largest benefit from the additional view predictions. 

While such simple strategy as using single-view alignment bridges the gap from single-view prediction to the optimal possible, the oracle row shows substantial remaining headroom (avg.\ $\text{CD}_n$: $0.025$ vs $\text{CD}_n$: $0.028$), suggesting that improved selection strategies, potentially leveraging learned scoring functions or multi-view consistency checks, could yield further gains. \looseness=-1

\begin{figure*}[!t]
    \centering
    \includegraphics[width=\textwidth]{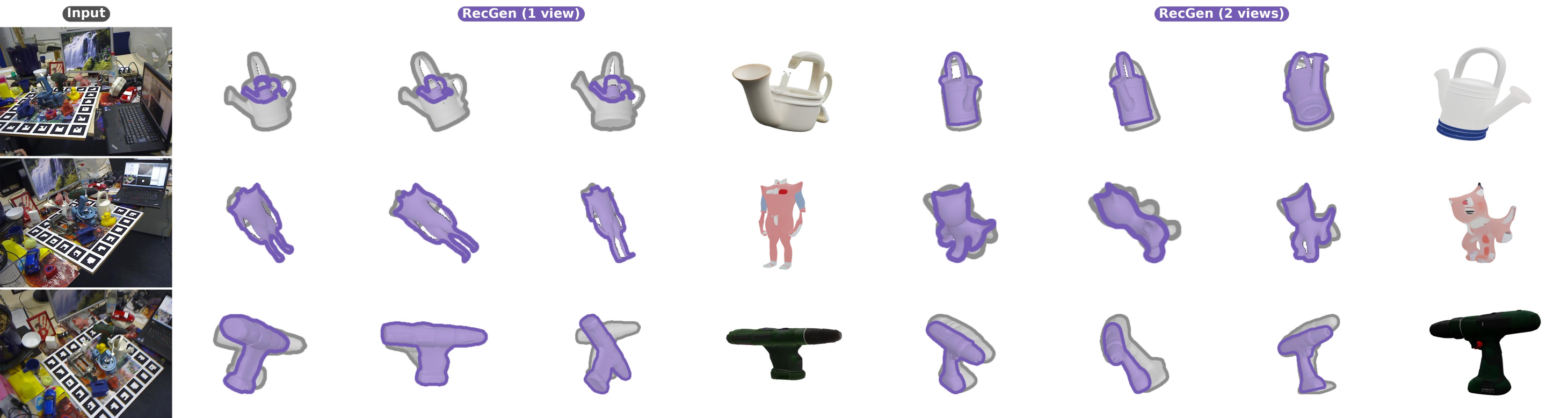}
    \caption{\textbf{Single-view vs.\ two-view reconstruction on LMO.} Each row shows one object: input image, three novel-view overlays (ground-truth in grey, prediction in purple) and a Gaussian Splatting render for the single-view (left group) and two-view (right group) settings. The second view reduces shape ambiguity, yielding reconstructions that more closely match the ground truth in scale (first row), appearance (second row) and rotations (third row).}
    \label{fig:multiview_pose_selection}
\end{figure*}

\Fig{fig:multiview_pose_selection} provides qualitative examples from LMO illustrating how the second view improves reconstruction quality. In each row, we show the input image alongside novel-view overlays of the ground-truth mesh (grey) and the \method posed shape prediction (purple), as well as a Gaussian Splatting render. With only a single view, the predicted shape often deviates from the ground truth in unseen regions. Adding a second view consistently improves the alignment, producing tighter overlaps with the ground-truth geometry and more detailed appearance.

\subsection{Multi-sample Generation Evaluation and Alignment-based Sample Selection}
\label{sec:multisample-selection}

\begin{table}[t]
\centering
\small
\setlength{\tabcolsep}{5pt}
\renewcommand{\arraystretch}{1.15}
\caption{Effect of multi-sample generation selection on HB (538 samples, 5 seeds). \emph{Single seed} reports the mean $\pm$ std across 5 independent generations. \emph{Pointmap alignment} selects, for each instance, the seed whose mesh best aligns with the input view's pointmap in metric camera space. \emph{Oracle} selects the seed with the lowest GT ADD-SB. Lower is better for both metrics.}
\begin{tabular}{lcc}
\toprule
Method & $\text{CD}_n$ $\downarrow$ & ADD-SB $\downarrow$ \\
\midrule
\method~(single seed) & $0.031 \pm 0.001$ & $0.048 \pm 0.001$ \\
\method~(pointmap alignment) & 0.029 & 0.043 \\
\cmidrule(lr){1-3}
\color{gray} \method~(oracle, best of 5) & \color{gray} 0.023 & \color{gray} 0.037 \\
\bottomrule
\end{tabular}
\label{tab:multisample_selection}
\end{table}

Since \method's reconstruction pipeline involves a stochastic denoising process, running multiple generations with different random seeds produces diverse shape and pose predictions for the same input. We investigate whether selecting among multiple generations can improve results, analogously to the multi-view pose selection above.

We evaluate on HB using 5 independent seeds. For each seed, we obtain a full reconstruction with associated metrics. We consider two selection strategies: (1)~\emph{pointmap alignment}, which uses the same single-view metric-space alignment score as in the multi-view setting to pick the best generation and could be applied during inference, and (2)~\emph{oracle}, which selects the seed with the lowest GT $\text{CD}_n$ (to show if one of many generation hypotheses from partial input information is close to GT).

\Tab{tab:multisample_selection} reports the results. The single-seed baseline averages $\text{CD}_n = 0.031 \pm 0.001$ and ADD-SB $= 0.048 \pm 0.001$ across seeds, showing low variance between runs. Pointmap alignment selection improves to $\text{CD}_n = 0.029$ and ADD-SB $= 0.043$, capturing a portion of the oracle gap ($\text{CD}_n = 0.023$, ADD-SB $= 0.037$). The oracle best-of-5 result represents a 26\% improvement in $\text{CD}_n$ and 23\% in ADD-SB over a single seed, demonstrating substantial diversity across generations containing samples that are much closer to GT mesh. However, as there is a substantial gap to the optimal selection, more sophisticated selection mechanisms or using the 2-view \method are needed for effective selection between the generated samples.

\section{Inference Efficiency}
\label{sec:efficiency}

A key advantage of \method's architecture design is that pointmaps and masks are fused additively with DINOv2 features of the inputs into a shared representation, rather than maintained as separate representations. This allows \method to be more efficient in terms of the memory and compute speed and allows for extensions to a multi-view version of the \method.  To confirm this, we measure wall-clock time and peak GPU memory (total process usage for 10 objects from the HB dataset, excluding model loading and post-processing (mesh export, rendering) in comparison to the SAM3D baseline. \Tab{tab:efficiency} compares the inference cost of \method and SAM3D on a single NVIDIA A100-SXM4-80GB GPU. \method is $1.8\times$ faster and requires $1.6\times$ less GPU memory than SAM3D.

\begin{table}[t]
    \centering
    \caption{\textbf{Inference efficiency comparison.} Measured on a single NVIDIA A100-SXM4-80GB, averaged over 10 HB samples. \emph{Allocated} is peak PyTorch tensor memory; \emph{Total} is full process GPU usage (\texttt{nvidia-smi}).}
    \label{tab:efficiency}
    \setlength{\tabcolsep}{5pt}
    \begin{tabular}{lccc}
        \toprule
        Method & \makecell{Allocated\\Memory (GB)} $\downarrow$ & \makecell{Total GPU\\Memory (GB)} $\downarrow$ & \makecell{Inference\\Time (s)} $\downarrow$ \\
        \midrule
        SAM3D~\cite{chen2025sam}       & $17.8 \pm 0.4$ & $22.0 \pm 1.4$ & $13.0 \pm 1.4$ \\
        \method                        & $10.4 \pm 0.5$ & $14.1 \pm 1.6$ & $\phantom{0}7.3 \pm 0.2$ \\
        \bottomrule
    \end{tabular}
\end{table}

\section{Limitations and Future Work}
\label{sec:limitations}
While \method demonstrates strong performance across diverse benchmarks, several limitations remain. First, \method assumes access to accurate object segmentation masks. When masks are imprecise---for example, including background regions---background depth values can bleed into the object pointmap, corrupting the geometric conditioning signal and degrading both pose and shape estimation. 
Second, the quality of generated textures and shapes is inherently bounded by the capacity of the underlying TRELLIS VAE for representing assets~\cite{xiang2025structured}. While our pose-conditioned appearance generation ensures correct texture orientation, fine-grained surface and geometry details are sometimes lost during the latent encoding and Gaussian Splatting-based decoding pipeline. Incorporating higher-capacity decoders, such as~\cite{xiang2025trellis2}, could further improve appearance fidelity. Third, \method's inference speed currently limits its applicability to real-time applications. With 50 denoising steps per stage across two generative stages, plus mesh extraction and texture baking, the full pipeline requires several seconds per object on a single GPU. While \method is already $1.8\times$ faster than SAM3D (\app{sec:efficiency}), this remains far from the real-time requirements of interactive robotic manipulation or augmented reality applications.

\paragraph{Future work.}
Several promising directions emerge from the current work. A natural extension is to generate not only geometric and visual properties but also \emph{physical parameters} such as mass, friction coefficients, collision geometries, and articulation joint types. Enriching the reconstructed assets with these properties would produce simulation-ready digital twins that can be directly imported into physics engines, significantly improving the utility of \method for real-to-sim transfer in robot learning pipelines. Another compelling direction is extending \method to the \emph{dynamic} setting: given video observations, the model could jointly reconstruct objects and their motion trajectories, enabling scene understanding that captures temporal evolution rather than a single static snapshot. Finally, addressing the limitations outlined above represents important future work: developing end-to-end pipelines that jointly perform segmentation and reconstruction, adopting more expressive appearance decoders from recent advances in 3D generation~\cite{xiang2025structured}, and exploring distillation or few-step denoising strategies to bring inference times closer to real-time operation.

\section{Detailed Baseline Description}
\label{sec:dataset-description}
The problem of simultaneously reconstructing objects and their parts is related to three research directions in recent literature: model-free pose estimation~\cite{lee2025any6d}, scene completion~\cite{agarwal2024scenecomplete}, and single-image 3D reconstruction~\cite{chen2025sam}. 

\paragraph{Model-free pose estimation.}
In model-free pose estimation, the goal is to estimate the pose of an object in an image (the query) given a reference view of the same object (the anchor). In several approaches, such as Any6D~\cite{lee2025any6d} and OneViewManyWords~\cite{geng2025one}, the target view may coincide with the reference view without affecting the method. For this reason, we selected Any6D as a baseline for model-free pose estimation.
Any6D generates a mesh using InstantMesh~\cite{xu2024instantmesh}. After an initial coarse alignment, it iteratively refines the pose and scale of the generated object to align it with the anchor image. At the end of the process, the mesh is scaled and the object pose is estimated. In the original model-free pose estimation setting, this mesh is then used to estimate the pose in the query image. Since our setup uses only a single image, we directly evaluate the mesh produced from the anchor view. Finally, we also experiment with replacing InstantMesh with TRELLIS~\cite{xiang2025structured}.

\paragraph{Scene completion.}
The objective of scene completion is to reconstruct complete object meshes or occupancy grids from a single-view RGB-D input~\cite{iwase2025zerograsp}. When applied to open-set scenarios, these methods become closely related to the problem of simultaneous object reconstruction and pose estimation. Among them, we selected SceneComplete~\cite{agarwal2024scenecomplete} as a baseline method.
SceneComplete proposes a modular architecture. The method first inpaints occluded objects by conditioning on a prompt produced by a vision-language model (VLM) and an estimate of the occluded region. InstantMesh~\cite{xu2024instantmesh} is then used to reconstruct the object from the inpainted image, while the scale is estimated using DINO features extracted from the rendered mesh. Finally, FoundationPose~\cite{wen2024foundationpose} aligns the reconstructed mesh with the RGB-D observation.
In the original work, the authors fine-tuned the inpainting module using LoRA to improve completion in the image plane. However, since the corresponding weights are not publicly available, we used the original pretrained weights for this module. To facilitate the inpainting process, we provided the ground-truth occlusion mask as input.

\paragraph{Single Image 3D reconstruction.}
The approach closest to \method is SAM3D~\cite{chen2025sam}, which simultaneously reconstructs objects and estimates their poses. SAM3D is proposed as a foundation model for 3D reconstruction, as it can generate aligned object meshes in an end-to-end manner. The method takes as input an RGB image, a segmentation mask of the target object, and optionally a point map of the scene. If the point map is not provided, it is estimated from the RGB image.
The model is a two-stage diffusion architecture. In the first stage, the model predicts the object pose, scale, and structured latents~\cite{xiang2025structured}. In the second stage, it generates the object mesh and Gaussian representations conditioned on the structured latents.
In our experiments, we provide the metric depth image to SAM3D to ensure a fair comparison.

\section{Detailed Datasets Description}
\label{sec:datasets-appendix}

\subsection{Training Datasets}
\begin{figure*}[!t]
    \centering
    \includegraphics[width=\linewidth]{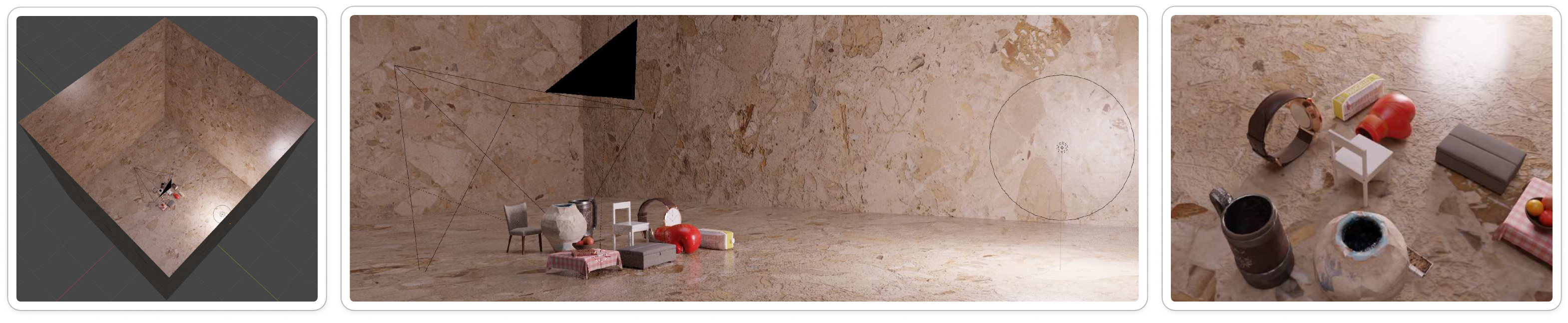}
    \caption{\textbf{Training dataset sample environment.} Example indoor rendering setup used for data generation. A primary object is placed on a table-like support surface and surrounded by 3--10 distractor objects. Materials, textures, and lighting are randomized to increase visual diversity. Scenes are rendered in BlenderProc with stereo camera views sampled around the main object.}
    \label{fig:dataset_setup}
\end{figure*}

Each object gets a mini indoor scene constructed from our asset pool, which comprises 198K high-quality 3D assets collected from 6 public object and part datasets: Objaverse-XL, ABO, HSSD, PhysXNet, PartNext, and PartNet-Mobility. Object-centric datasets (Objaverse-XL, ABO, HSSD) are used to create compositional tabletop scenes where 3-10 randomly selected distractor objects from the same source dataset are placed around the main object to induce natural occlusions and complex depth relationships. For part-centric datasets (PhysXNet, PartNext, PartNet-Mobility), scenes contain a single object due to significant self-occlusion among articulated or fine-grained parts.

Scenes are rendered using BlenderProc~\cite{denninger2019blenderproc}, with randomized indoor layouts, textures, materials, and lighting configurations to enhance visual diversity and realism. A sample indoor setup used for rendering is shown in~\fig{fig:dataset_setup}, illustrating the table-like support surface, surrounding distractors, and lighting arrangement. For each scene, twenty stereo camera views are sampled from varying azimuth, elevation, and distance around the primary object, producing a diverse set of viewpoints. In total, the dataset contains 198K scenes and 3.2M synthetically generated image pairs with and without occlusions.

\begin{figure*}[!t]
    \centering
    \includegraphics[width=\linewidth]{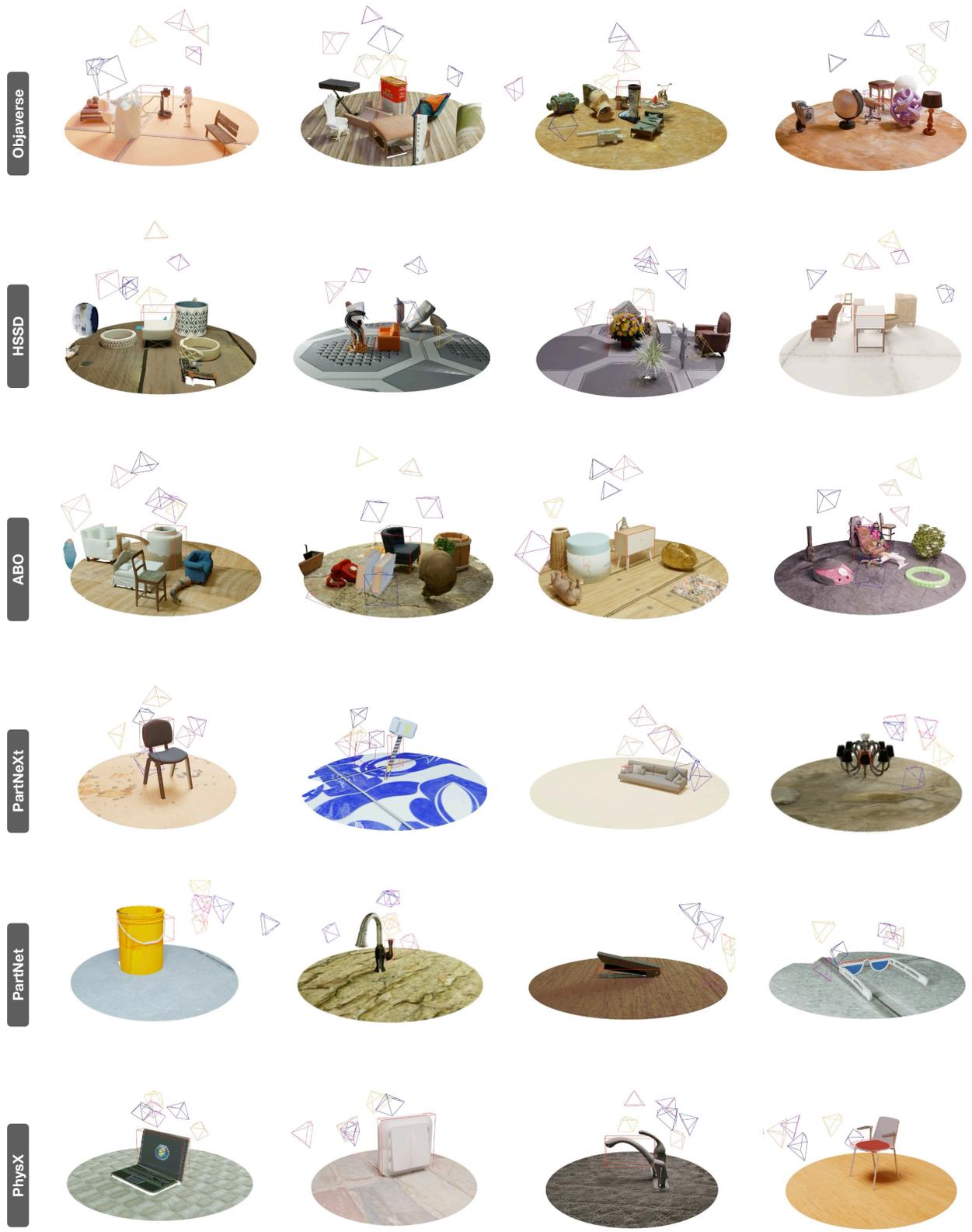}
    \caption{\textbf{\method training dataset samples.} Additional examples of 3D assets from our training dataset, including compositional scenes with objects from Objaverse-XL, ABO, HSSD, and parts in object scenes the part-based datasets from PhysXNet, PartNext, and PartNet-Mobility. }
    \label{fig:dataset_samples_extended}
\end{figure*}

For every rendered view, we provide RGB images, depth maps, stereo depth, semantic and instance segmentation masks, amodal masks, ground-truth 6D object poses, and full camera metadata. For appearance generation training, subsets from PartNet-Mobility and PhysXNet are excluded due to lower texture quality. Example samples from the different datasets are shown in \fig{fig:dataset_samples_extended}.

\subsection{Object-based Evaluation Datasets}
\label{sec:object-based-eval}

The experiments were conducted on four object-centric datasets: Linemod Occluded (LM-O)~\cite{brachmann2014learning}, NVIDIA Household Objects for Pose Estimation (HOPE) \cite{tyree2022hope}, ReOcS~\cite{iwase2025zerograsp}, and HomebrewedDB (HB)~\cite{kaskman2019homebreweddb}. We selected these datasets to capture a variety of object types, cameras, and occlusion levels. For each dataset, we sampled a random subset of frames. We use standard $3\times3$ mask erosion to avoid misalignment between the depth map and the masks at the borders.

\paragraph{LM-O.}
This dataset contains 8 objects with significant occlusions, providing a standard benchmark for pose estimation under partial visibility. The RGB-D images were captured using a structured-light sensor from the Kinect v1 / PrimeSense family. From this dataset, we randomly sampled 142 frames.

\paragraph{HOPE.}
It includes 28 toy grocery objects captured in 50 scenes across 10 household and office environments, with up to five lighting variations and varying levels of occlusion. The RGB-D data were acquired using an Intel RealSense D415 camera, a stereo-based depth sensor delivering synchronized high-resolution color and depth streams. We selected this dataset for its lighting diversity and the presence of objects that are symmetric in shape but asymmetric in texture. From this dataset, we sampled 506 frames.

\paragraph{HB.}
It comprises 33 diverse objects (17 toy, 8 household, and 8 industry-relevant) recorded in 13 scenes with varying levels of complexity. We sampled 538 frames from the Kinect subset to include time-of-flight (ToF) camera data.

\paragraph{ReOcS.}
It provides 3D shape and pose annotations for 22 unseen objects, along with high-quality depth maps generated via learning-based stereo matching. The dataset is divided into three splits according to occlusion levels. We sampled 314 frames from the normal split, which contains balanced occlusions.

\subsection{Part-based Evaluation Dataset}
\label{sec:part-based-eval}

In the real-to-sim domain, the ability to decompose objects into components is fundamental for creating realistic simulations that are reliable and useful for robot learning~\cite{kerr2025robot,learticulate,yu2025real2render2realscalingrobotdata}. Therefore, we believe that benchmarking the accuracy of methods that jointly perform reconstruction and pose estimation for object parts is crucial to understanding how reliable these approaches are for real-to-sim applications.
A desirable dataset for part-based pose estimation should include RGB-D images captured from multiple viewpoints, camera parameters, part meshes, part poses, and realistic object arrangements. To the best of our knowledge, none of the existing datasets that provide ground-truth meshes possess all these characteristics.
Since several works have demonstrated reliable sim-to-real transfer from scenes rendered with IsaacSim~\cite{singh2025synthetica, han2025re, dowdy2025isaac, yu2024orbit, wen2024foundationpose}, we decided to address this dataset shortage by leveraging this simulator.
We started from the ArtVIP~\cite{jin2025artvip} articulation dataset, which contains six predefined scenes. We extended these scenes with additional articulated objects while preserving their realistic structure. From these scenes, we rendered views around the objects and organized the resulting data in the BoP format.

The following paragraphs provide further details on the motivations behind the dataset and its construction.

\paragraph{Object part estimation in real-to-sim.}
Estimating the shape and pose of object parts is fundamental for real-to-sim pipelines and robot learning. In particular, accurate part estimation is a key step in constructing models of articulated objects.~\cite{artykov2025articulated,liu2023paris,guo2025articulatedgs,lin2025splart,kerr2025robot,learticulate}. When both the parts and their motions are correctly estimated, the resulting articulated models can be used to learn reliable manipulation policies.~\cite{learticulate,kerr2025robot}
Recent advances in automatic real2sim and real2sim2real pipelines further highlight this need.~\cite{torne2024reconciling,yu2025real2render2realscalingrobotdata, barcellona2025dream}. For instance, RialTo~\cite{torne2024reconciling} introduces a graphical interface for manual object part annotation. In DreMa~\cite{barcellona2025dream} the robot parts are reconstructed starting from segmentation masks. Similarly, Real2Render2Real~\cite{yu2025real2render2realscalingrobotdata} uses segmentation to extract object parts, but additionally leverages videos of object motion to estimate their dynamics.

\paragraph{Dataset Generation.} 
ArtVIP contains six scenes: children\_room, dining\_room, kitchen, \\
kitchen\_with\_parlor, large\_living\_room, and small\_living\_room. For each scene, we created an additional replica containing more articulated objects defined in the ArtVIP dataset, resulting in a total of 12 scenes.
For each object, we rendered 40 RGB-D images and visible segmentation masks using cameras uniformly sampled on a hemisphere around the object. We used a radius of 1.7 meters for all objects, except for those in the kitchen scene, where we used a radius of 1.2 meters because the objects are closer to each other. For objects that would otherwise fall outside the image frame, we increased the camera radius accordingly. From the 40 views, we discarded small masks or objects that were completely occluded.
For each object, we extracted its constituent parts. We define parts as components of an object mesh whose motion depends on the object's structure. Parts can either be fixed elements (e.g., the legs of a chair) or movable components that change the object's state (e.g., articulated elements such as drawers). In addition to the rendered part masks, we also generated the full object mask.
Finally, we extracted each part mesh in the world coordinate frame of the simulation, translated it to the origin, and computed its pose in each camera frame, organizing the resulting data in the BoP format.

\begin{figure*}[!t]
    \centering
    \includegraphics[width=0.9\linewidth]{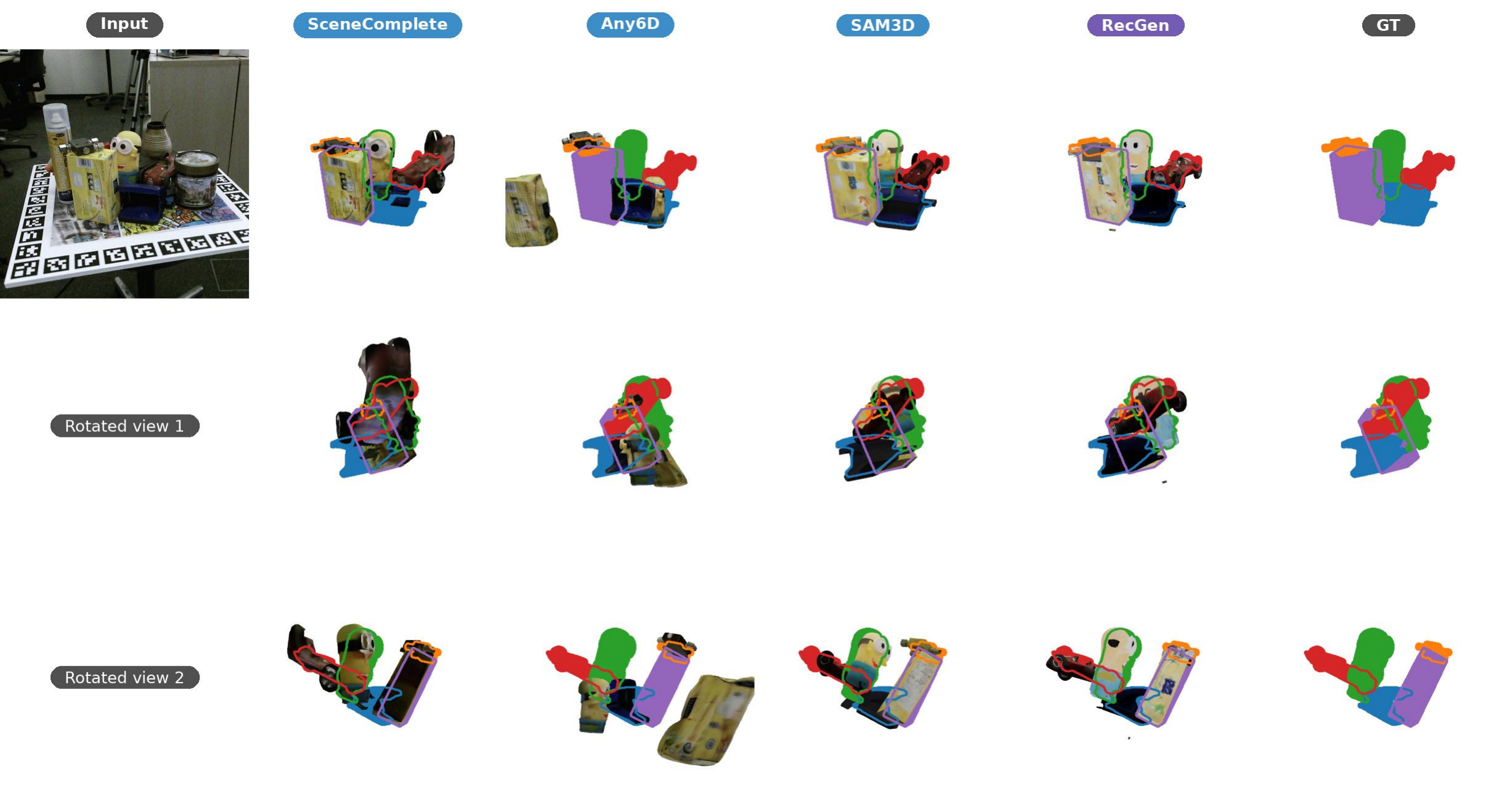} \\[2pt]
    \includegraphics[width=0.9\linewidth]{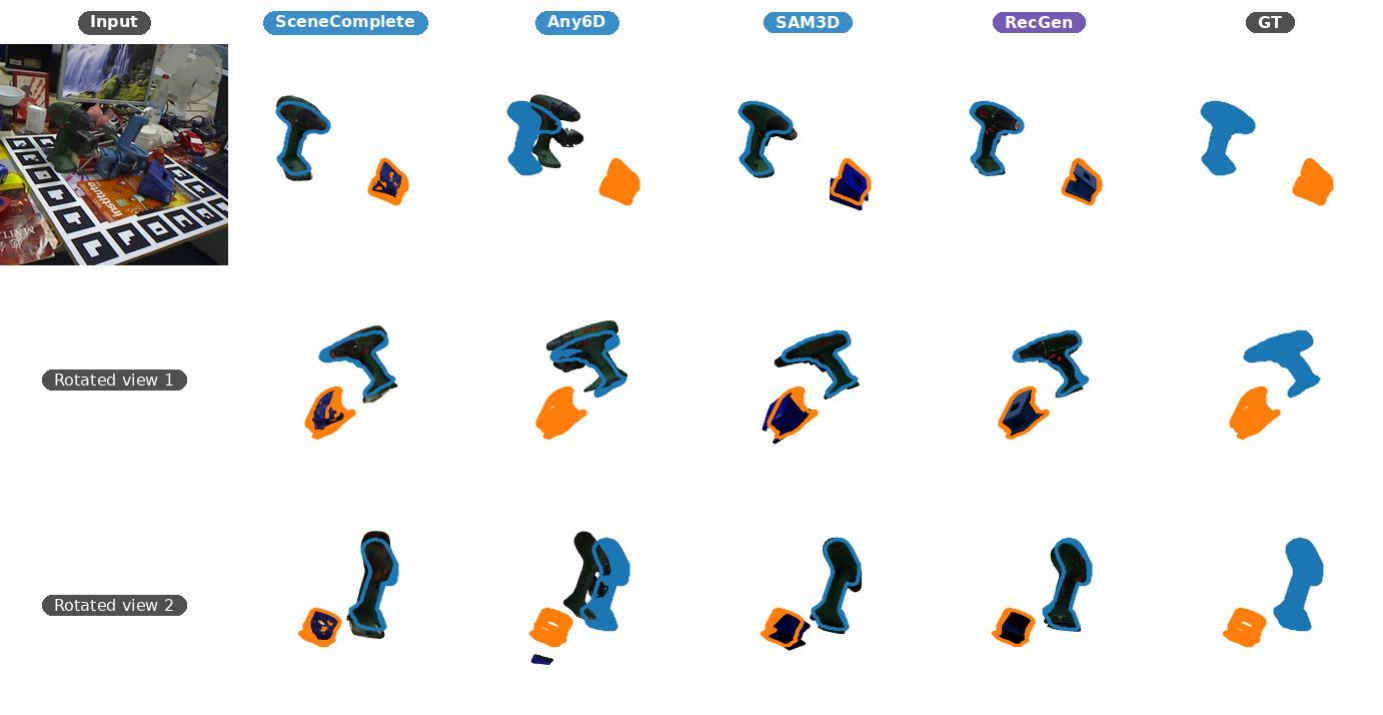} \\[2pt]
    \includegraphics[width=0.9\linewidth]{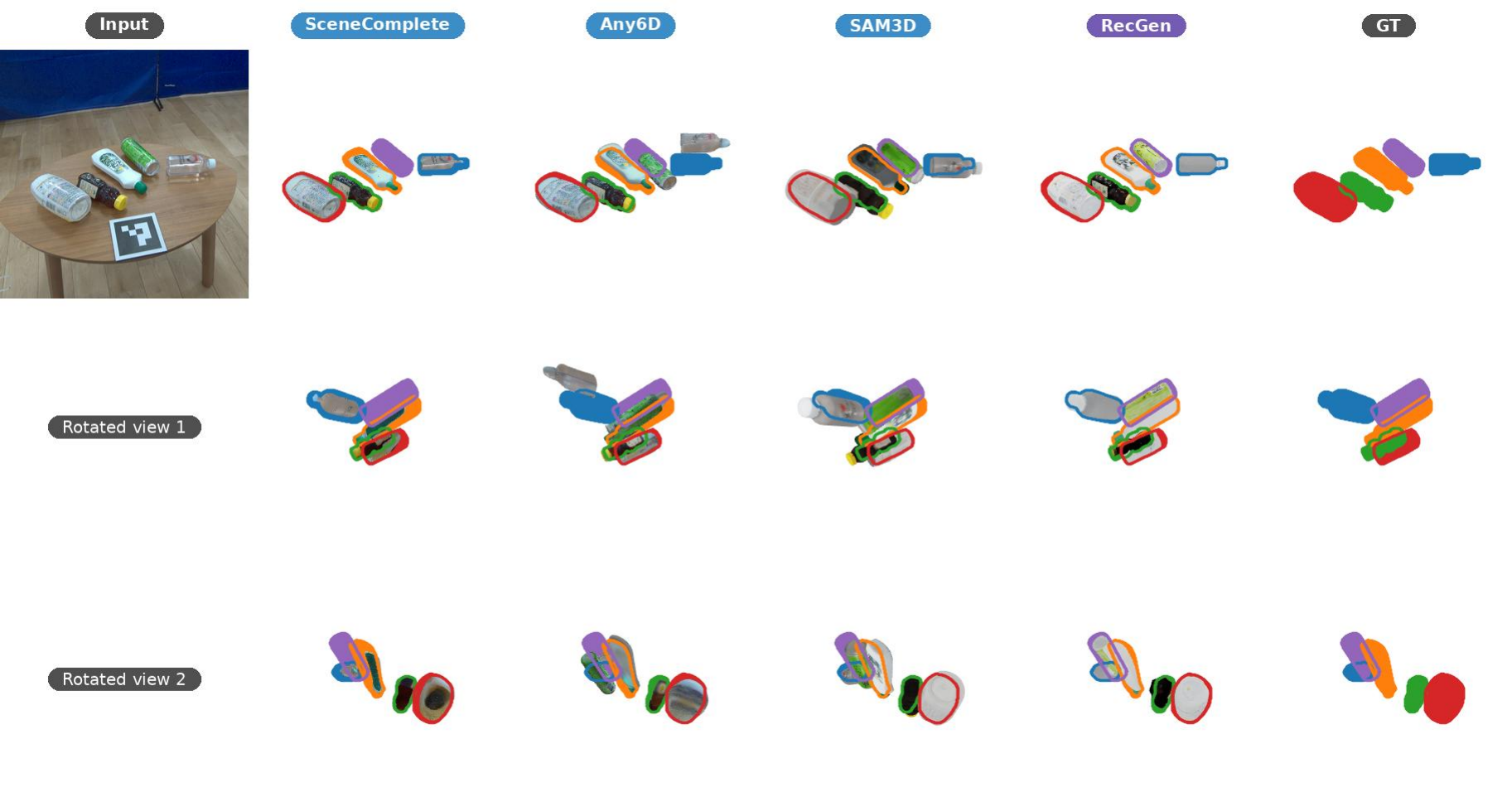}
    \caption{\textbf{Qualitative comparison with baselines.} We compare \method with SceneComplete, Any6D, and SAM3D on scenes from HB, LMO, and ReOcS datasets. Each row group shows the input image, reconstructions from each method, and the ground truth (GT) from three viewpoints.}
    \label{fig:comparison_appendix}
\end{figure*}

\paragraph{Evaluation.}
From the generated dataset, we randomly sampled up to 4 views per part to construct the test set. At this stage, we initially selected 284 objects. We then manually filtered the objects to avoid oversampling parts that are widely represented in the dataset. After this process, the number of test objects was reduced to 262, corresponding to a total of 924 test frames. From this sample, we further randomly selected 500 frames for which a second view was available, enabling multi-view evaluation. The evaluation procedure follows the same protocol used for the other datasets.

\section{Additional Qualitative Comparison}

\subsection{Object-based Reconstruction}
\label{sec:object-based-appendix}

\Fig{fig:comparison_appendix} presents additional qualitative comparisons between \method and all baselines on scenes from HB, LMO, and ReOcS datasets, showing reconstructions from three viewpoints.

\subsection{Part-based Reconstruction}
\label{sec:part-based-appendix}

\begin{figure*}[!t]
    \centering
    \includegraphics[width=\textwidth]{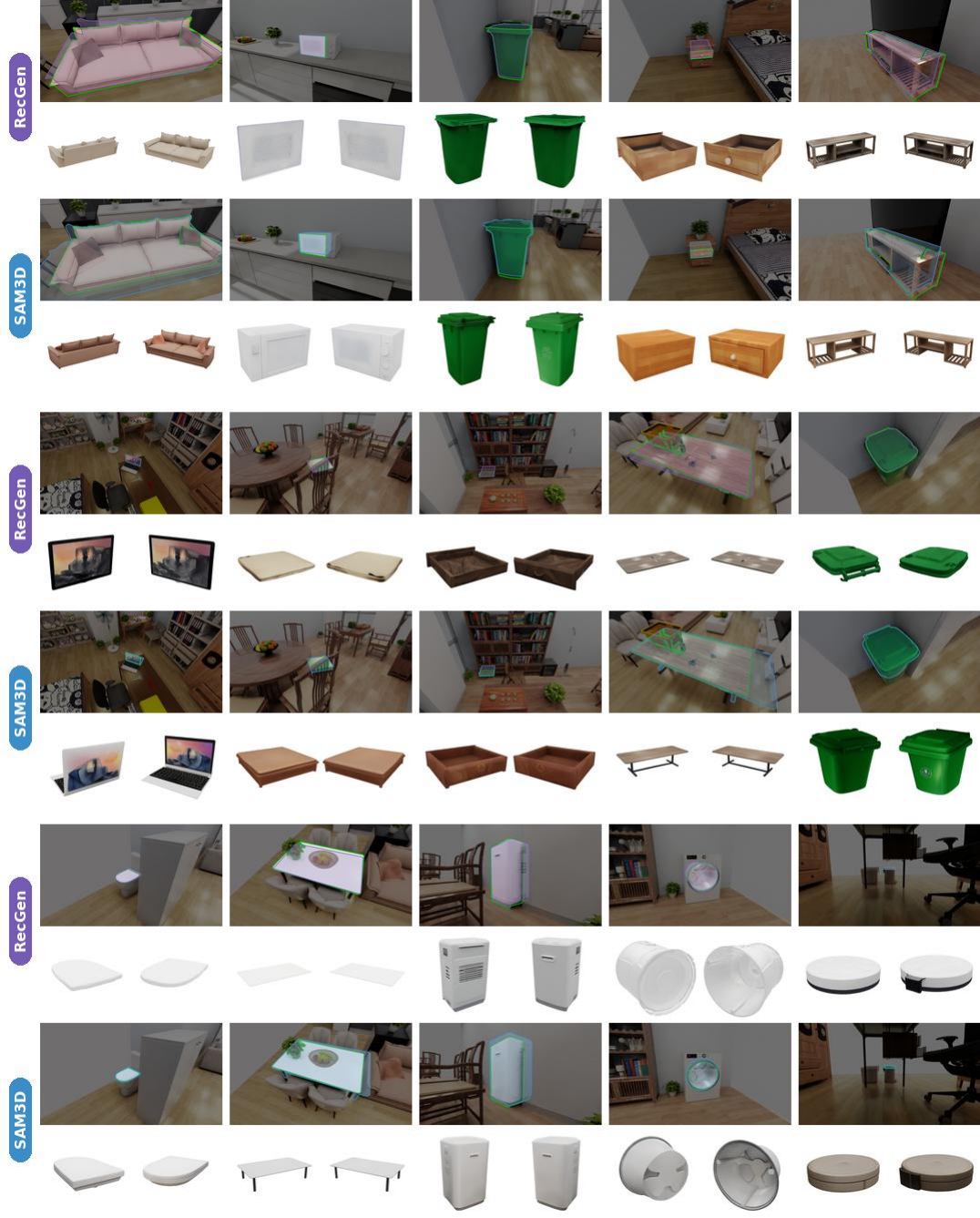}
    \caption{\textbf{Part-based reconstruction: \method vs.\ SAM3D on ArtVIP.} Each column shows one articulated part. For each method: scene overlay with GT mask contour (green) and predicted mesh (purple/blue), plus two novel-view GS renders. \method produces more accurate part geometry across diverse categories.}
    \label{fig:part-based-comparison}
\end{figure*}

\Fig{fig:part-based-comparison} presents a qualitative comparison between \method and SAM3D on part reconstruction from the ArtVIP dataset. Across diverse object parts categories (such as drawers, doors, lids, and appliance parts) \method generates reconstructions that more faithfully capture the part geometry, particularly for thin structures and parts under partial self-occlusion.

\FloatBarrier

\end{document}